%% file: main.tex
\pdfoutput=1

\documentclass[11pt]{article}

\usepackage[]{emnlp2023-latex/EMNLP2023}

\usepackage{times}
\usepackage{latexsym}

\usepackage[T1]{fontenc}

\usepackage[utf8]{inputenc}

\usepackage{microtype}

\usepackage{inconsolata}

\usepackage{tablefootnote}
\usepackage{indentfirst}
\usepackage{array}
\usepackage{graphicx}
\usepackage{adjustbox}
\usepackage{amsmath,amssymb,amsthm,bm}
\usepackage{caption}
\usepackage{multirow}
\usepackage{booktabs}
\usepackage{listings}
\usepackage{subcaption}
\usepackage{bbding}
\usepackage{cleveref}
\usepackage{xfrac}
\usepackage{svg}
\usepackage{float}
\usepackage{xcolor, soul}

\usepackage{paralist}

\usepackage{xspace}
\usepackage{cleveref}

\usepackage{tikz}

\input{Definitions}

\newcommand{\dialgen}{\textsc{DialGen}\xspace}
\newcommand{\aic}{\textsc{AIC}\xspace}

\newcommand{\syntaic}{\textsc{DialGen-AIC}\xspace}
\newcommand{\Qavg}{$CB_{avg}$\xspace}
\newcommand{\Qu}{$CB_{1}$\xspace} 
\newcommand{\Qd}{$CB_{2}$\xspace}
\newcommand{\Qt}{$CB_{3}$\xspace} 
\newcommand{\Qc}{$CB_{4}$\xspace} 

\newcommand{\ensuretext}[1]{#1}

\newcommand{\arkcomment}[3]{\ensuretext{\textcolor{#3}{[#1 #2]}}}
\renewcommand{\arkcomment}[3]{}  

\definecolor{green}{RGB}{0,128,0}
\definecolor{lightgray}{RGB}{211, 211, 211}
\definecolor{pink}{RGB}{255,0,255}
\definecolor{red}{RGB}{219,68,55}
\definecolor{yellow}{RGB}{244,180,0}
\definecolor{blue}{RGB}{66,133,244}
\definecolor{lightyellow}{RGB}{255,255,0}
\sethlcolor{lightyellow}

\newcolumntype{C}[1]{>{\centering\arraybackslash}p{#1}}
\newcolumntype{R}[1]{>{\raggedleft\arraybackslash}p{#1}}

\title{Does Collaborative Human--LM Dialogue Generation \\ Help Information Extraction from Human Dialogues?}

\DeclareSymbolFont{extraup}{U}{zavm}{m}{n}
\DeclareMathSymbol{\varheart}{\mathalpha}{extraup}{86}
\DeclareMathSymbol{\vardiamond}{\mathalpha}{extraup}{87}

\newcommand\uw{$^\spadesuit$}
\newcommand\msr{$^{\heartsuit}$}
\newcommand\as{$^{\varheart}$}
\newcommand\uhk{$^{\clubsuit}$}
\newcommand\ai{$^{\diamondsuit}$}
\newcommand\aspace{\hspace{.75em}}
\newcommand\first{$^\ast$}

 \author{
     Bo-Ru Lu\uw\thanks{\first Equal contribution.}\aspace
     Nikita Haduong\uw\first\aspace
     Chia-Hsuan Lee\uw\aspace
     Zeqiu Wu\uw\aspace
     \textbf{Hao Cheng}\msr\aspace \\
     \textbf{Paul Koester}\as\aspace
     \textbf{Jean Utke}\as\aspace
     \textbf{Tao Yu}\uhk\aspace
     \textbf{Noah A. Smith}\uw\ai\aspace
     \textbf{Mari Ostendorf}\uw\aspace\\
     \uw{}University of Washington \aspace
     \msr{}Microsoft Research \aspace \\
     \as{}Allstate \aspace
     \uhk{}University of Hong Kong \aspace
     \ai{}Allen Institute for AI\\
     {\tt \{roylu,chiahlee,zeqiuwu1,ostendor\}@washington.edu
     chehao@microsoft.com}\\
     {\tt 
     \{pkoes,jutke\}@allstate.com
     tyu@cs.hku.hk
     \{qu,nasmith\}@cs.washington.edu}
}

\begin{document}
\maketitle
\begin{abstract}
\input{sections/0_abstract}
\end{abstract}

\input{sections/1_intro}

\input{sections/2_framework_new}
\input{sections/3_def_and_eval}
\input{sections/4_dataset}
\input{sections/5_experiments}
\input{sections/5.1_results}
\input{sections/6_analysis}
\input{sections/8_conclusion}
\input{sections/9_limitations}

\bibliography{custom}
\bibliographystyle{emnlp2023-latex/acl_natbib}

\appendix
\include{sections/appendix}

\end{document}

%% file: sections/0_abstract.tex
The capabilities of pretrained language models have opened opportunities to explore new application areas, but applications involving human-human interaction are limited by the fact that most data is protected from public release for privacy reasons. 
Problem-solving human dialogues in real applications can be much more complex than existing Wizard-of-Oz collections, preventing successful domain transfer. 
To support information extraction (IE) for a private call center dataset, we introduce a human-in-the-loop 
dialogue generation framework capable of synthesizing realistic dialogues. 
In IE experiments with auto insurance call center dialogues, we observe 25\% relative improvement in $F_1$ after augmenting a small set of real human conversations with synthetic data. 
We release code and our synthetic dataset to illustrate the complexity of real-world call center conversations and encourage development of complex dialogue datasets that are more representative of natural data.

%% file: sections/1_intro.tex
\section{Introduction} 
\input{figure_tex/teaser}

Rapid advances in natural language processing have driven interest in its use in a wide variety of domains.  However, applications involving human-human interaction, such as call center dialogues, have seen limited success. One reason is that natural problem-solving dialogues are not typically publicly available for privacy reasons, restricting opportunities for researchers to explore methods in advancing applications for these domains. Further, annotating private datasets can be expensive because of the need for in-house expertise, so training resources are limited. In this paper, we introduce a method to fill the data gap using synthetic data generated by a collaborative human--language model framework. Specifically, we experiment with a task of extracting information from auto insurance call center dialogues, using public synthetic data to improve performance on a private dataset.

Many available dialogue datasets are designed for training \emph{virtual agents}, collected using pairs of humans to perform a task \cite{budzianowski2018multiwoz,rastogi2020towards,chen2021action}.
Designing for human-machine interaction results in dialogues that lack the complexity of human-human dialogues. Additionally, human-only data collection can have limited content diversity, result in imbalanced training sets, and does not scale to more complex tasks, due to the high cost of employing domain experts \cite{geva-etal-2019-modeling,gururangan-etal-2018-annotation,qian-etal-2021-annotation}. 

To reduce data collection costs, researchers have explored using language models (LMs) to generate synthetic training data 
\cite{bao-etal-2023-synthetic,8621223,10.1162/tacl_a_00492,li-etal-2023-synthetic,auto-event-extraction-datagen, singan}.
Synthesized data can target long-tail phenomena \cite{10.1007/978-3-030-58526-6_41} and allow for public release of data that closely emulates real-world privacy-constrained domains, such as the medical domain~\citep{dataSynthesisPrivacyParkGan}. While LMs can follow instructions to generate text that closely resembles human writing, there can be challenges ensuring the data is diverse and not too simplistic \citep{synsys,stahlberg-kumar-2021-synthetic,liu2022wanli}. In addition, they still suffer from issues with incoherence and consistency \cite{clark-etal-2021-thats,dou-etal-2022-gpt}. To protect against LM errors, collaborative human-LM frameworks have been designed for tasks involving short dialogues and texts \cite{liu2022wanli,bonaldi-etal-2022-human}.
In contrast, we investigate using a human-in-the-loop framework to create \textbf{lengthy and complex dialogues}. 

Our work proposes a human-LM collaborative framework for dialogue generation (\dialgen) that leverages the scalability and creativity of generative models, yet retains controllability through humans. Human collaborators edit the synthesized dialogues, which we use to boost information extraction performance on real-world call center data. 

Many call center dialogues involve problem solving where customers provide information to an agent through question-answer pairs and clarifications that need to be interpreted in the context of the dialogue history. Our information extraction (IE) task is thus framed as an iterative information update after each agent-customer exchange, analogous to dialogue state tracking (DST) in task-oriented dialogues. 
However, unlike DST, the information extracted from each turn is collected to create a summary of the call rather than to generate a virtual agent's response or make an API call.
In addition, the state includes entities that are associated with attributes (slots) and values. 
To evaluate models on this IE task, we introduce entity-centric scoring methods that allow for partial matching of multiple and descriptive values.

We demonstrate the effectiveness of \dialgen by generating data in auto insurance calls, a domain with privacy restrictions preventing public release of natural calls, and performing information extraction. 
We work with a private dataset containing 34 dialogues with an average 197 utterances per dialogue and synthesize 235 dialogues with an average 46 utterances per dialogue. Experiments in our IE task show the additional synthetic data improves model performance by 25\% in the full $F_1$ score.

To summarize, our main contributions are: 
\begin{itemize}
    \item We design \dialgen, a collaborative human-LM framework for generating complex task-oriented dialogues in domains where privacy constraints have previously prevented data sharing with the research community. Synthetic data, training documentation, prompts, and interface code will be released.\footnote{\url{https://boru-roylu.github.io/DialGen}}
    \item We present \syntaic, a custom dataset designed to illustrate the complexity of real-world auto insurance call center data. While not intended as a benchmark, \syntaic aims to provide a demonstration for the complex nature of real conversations and the challenges faced in this domain, including linking information with different entities and tracking multiple values in a single slot.
    \item We propose an entity-centric scoring methodology that considers information links to different entities, allows for multiple slot values, and provides partial match scores for descriptive values. 
\end{itemize}

%% file: figure_tex/teaser.tex
\begin{figure}[ht]
    \centering    
    \includegraphics[width=\linewidth]{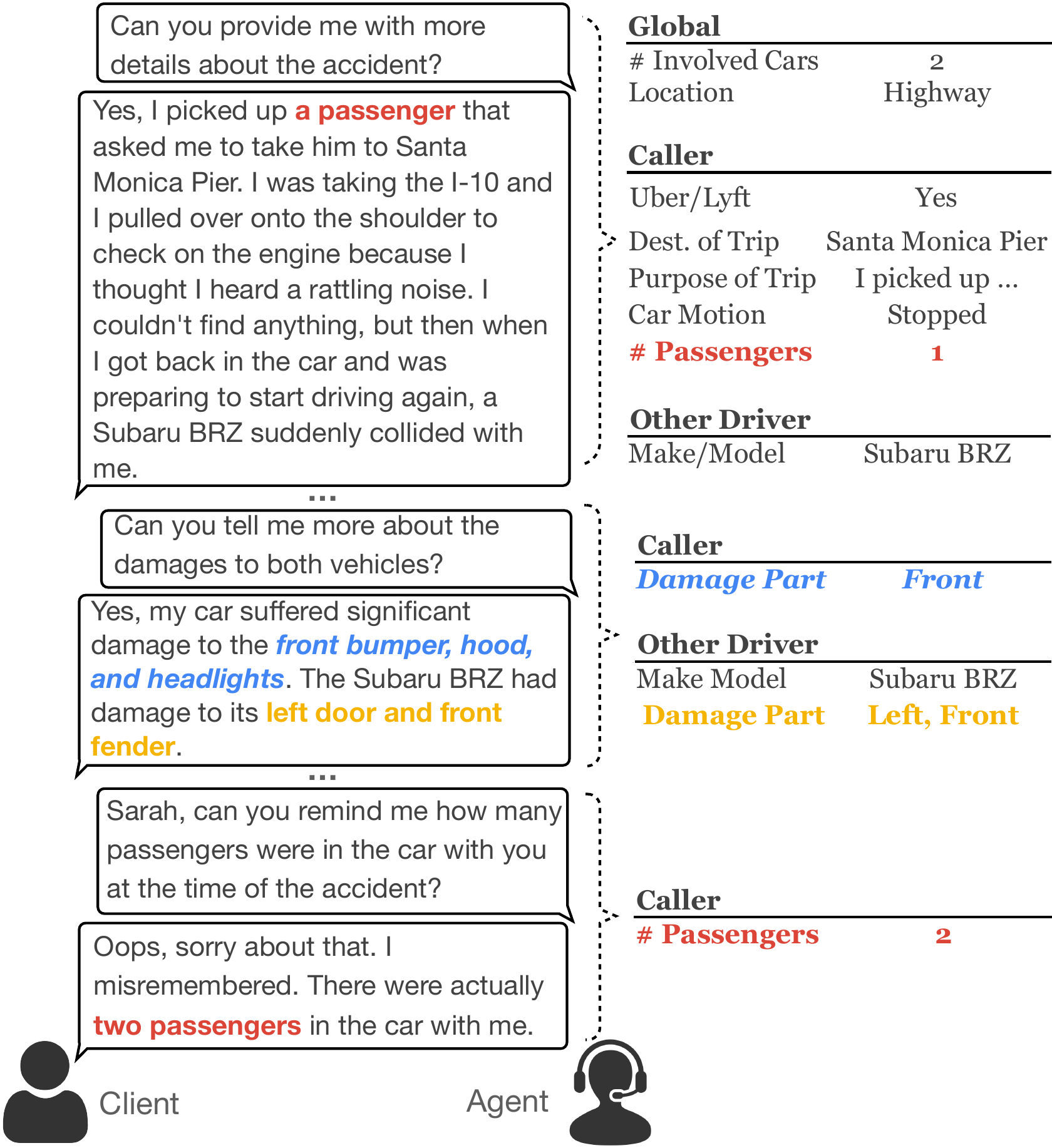}
    \caption{An illustrative snippet of our dialogue with entity-slot-value triples. \textcolor{yellow}{\bf{Yellow}} is the slot with multiple values. \textcolor{blue}{\textit{\textbf{Italic blue}}} and \textcolor{yellow}{\bf{yellow}} are the same slot (\textit{Damage Part}) with different entities (e.g., \textit{Caller} and \textit{Other Driver}). \textcolor{red}{\bf{Red}} is a slot with a value update.}
    \label{fig:teaser}
\end{figure}

%% file: sections/2_framework_new.tex
\input{figure_tex/framework}

\section{Dialogue Generation (\dialgen)}

As shown in \autoref{fig:pipeline}, our \dialgen framework is designed to generate schema-guided dialogues through human-LM collaboration.
An LM is selected as the backbone, then the data generation process begins with an initial task prompt consisting of natural language description for the desired dialogue (e.g., task description, desired slots, story, and personalities) and dialogue history.
During each iteration, the LM first proposes a candidate subdialogue based on the history (the initial task prompt and the generated conversation so far).
Human reviewers with sufficient domain knowledge then validate, edit, and annotate the generated subdialogue, before requesting a continuation via an updated prompt to the LM.
The reviewers can optionally augment the prompt with a specific instruction related to the desired dialogue flow.
This process repeats until the dialogue is complete. 
At a high level, the human-in-the-loop mechanism ensures that the resulting dialogues are coherent and consistent with the prompt, covering desired content and fulfilling style specifications from domain experts.
In the following, we describe each component of \dialgen in detail.

\subsection{Prompt for Dialogue Generation} 
The prompt for generating synthetic dialogues includes:
the task description, entity-slot-value triplets, story, personality and dialogue history.\footnote{An example of a full prompt is given in Appendix~\ref{appendix:dialogue-generation-prompt}.}

\paragraph{Task Description.} Similar to task descriptions given to humans in Wizard-of-Oz setups \citep{kelley1984iterative}, the template-based task description gives the information about dialogue participants and the task scenario for the conversation, such as having the LM role-play as a user calling to file a claim with an agent at an insurance company, e.g., ``\textit{Role play car accident claim call. One person is an agent Alice from a car insurance company and the other is the caller Bob who wants to file a claim.}''
    
\paragraph{Entity-slot-value Triplets.} We randomly sample entity-slot-value triples from the expert-authored ontology to steer the LM to generate required content in the dialogue, enabling precise covering of specific information, e.g., (\textit{Caller, Injury, Neck}).
    
\paragraph{Story.} \citet{kim2022soda} synthesize social dialogues from common sense knowledge triples by first using a social narrative to set up the scenario. We similarly use the randomly sampled triplets to generate a story with the LM before the dialogue generation. For example, the aforementioned entity-slot-value triple will be converted into the snippet of a story: ``\textit{The impact of the collision caused Bob’s car to spin around and come to a stop. He immediately felt a sharp pain in his neck and knew that something was wrong}.''

\paragraph{Personality.}
To enrich the diversity of callers, we randomly sample a personality from the predefined list (\autoref{tab:personality}) for each dialogue, e.g., ``\textit{Bob is feeling distressed or frustrated due to the accident and its consequences.}'' 
For the agent, we use the same personality for all dialogues, e.g., ``\textit{Alice is conversational, personable, patient, empathetic, sympathetic and professional.}''

\paragraph{Dialogue History.} The LM uses the full dialogue history to generate subdialogue turns that are consistent with the flow of the conversation. During the subdialogue generation process, we append completed subdialogues before generating the next subdialogue.
The initial dialogue history is always one exchange, e.g., 
``\textit{Alice: Hi, thank you for calling DialGen Insurance! This is Alice. How may I help you today?}'' followed by ``\textit{Bob: I am calling regarding a car accident.}''

\subsection{Subdialogue Generation}
The dialogue is generated iteratively where each subdialogue is revised and annotated by a reviewer. 

\paragraph{Human-in-the-loop Review.}
Subdialogues are individually revised by a human trained to correct common LM errors such as those described by~\citet{dou2021scarecrow}, verify that required information is present (the sampled triples), and edit the text to meet stylistic criteria (e.g., adjusting tone). The reviewer can either revise individual turns directly or instruct the LM to regenerate specified turns,
e.g., ``\textit{Have the caller correct earlier incorrect information}'' (more examples in \autoref{tab:instruction_prompts}). 
The LM may try to end the dialogue by including termination signals such as ``\textit{good bye}.'' If the LM ends the dialogue without covering the required triplets, the reviewer can delete and regenerate the turns.

\vspace{-2mm}
\paragraph{Annotation.}
Spans in the subdialogue that have information tuples associated with the task ontology are annotated by the human reviewer.  If a tuple in turn $t$ has a slot with the same referent and a different value than a previous turn, the reviewer is asked to resolve the duplication by indicating whether the new value is a correction \textsc{update}, \textsc{keep}, or additional detail to be concatenated with the previous value \textsc{concat}. After annotation, the review can choose to generate another subdialogue or accept the ending that the LM has proposed. This annotation step is optional and can be decoupled from the \dialgen framework depending on the target tasks or domains.

%% file: figure_tex/framework.tex
\begin{figure*}[htp!]
    \centering
    \includegraphics[width=0.95\textwidth]{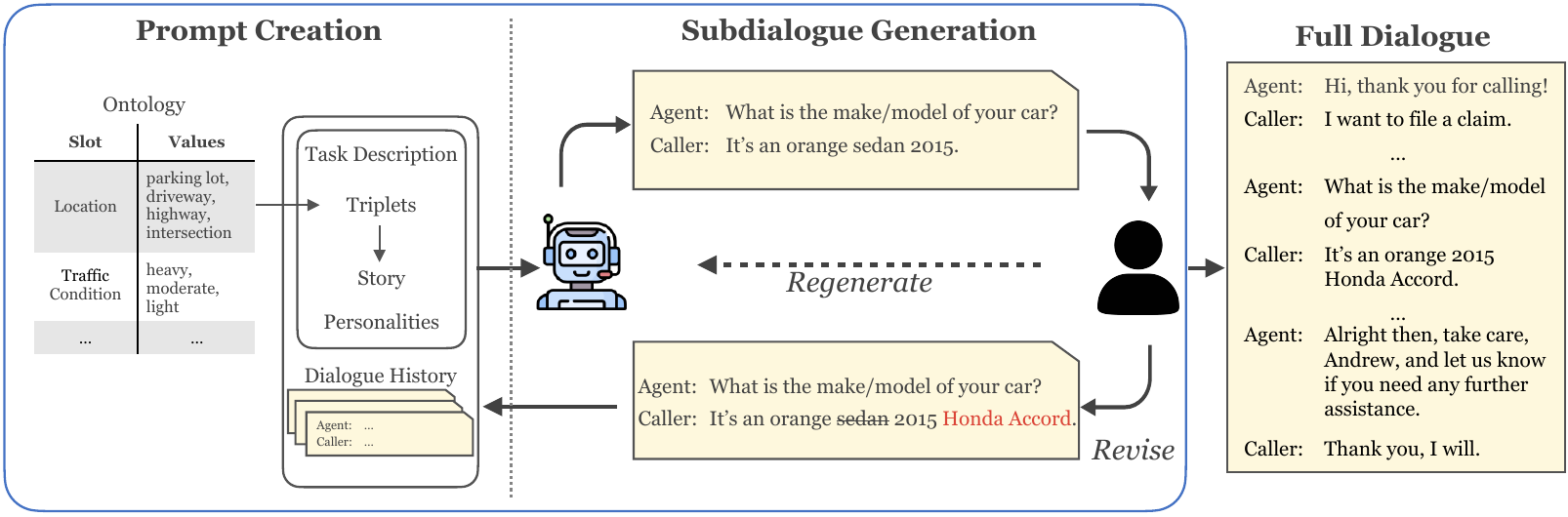}
    \caption{In the \dialgen framework, a language model (LM) and a human reviewer collaborate to generate a dialogue.
    First, a story is created by the LM, using randomly sampled entity-slot-value triplets from the ontology. Second, the LM generates a subdialogue, using a task description, triplets, story, personalities, and dialogue history. 
    The reviewer evaluates how the subdialogue fits with the task requirements and dialogue history. 
    If not satisfied, the reviewer can have the LM regenerate the subdialogue before revising it. 
    The revised subdialogue is added to the dialogue history for generating the next subdialogue. 
    This iterative process continues until the dialogue is complete.}
    \label{fig:pipeline}
\end{figure*}
\vspace{-5mm}

%% file: sections/3_def_and_eval.tex
\section{Problem Definition and Evaluation}
\label{sec:dst_extension}
Auto insurance call center dialogues involve customers working together with an agent to address an issue or submit a claim. As the conversation progresses, extracted information must be iteratively updated.
This updating process is similar to the concept of dialogue state tracking (DST) used in task-oriented dialogues.
However, unlike standard DST, the extracted information is used to summarize the call, not to make API calls or generate responses by a virtual agent.

\subsection{Problem Definition}
Extracted structured information is typically represented as a collection of tuples $\{ (s,v), s \in \mS \}$, where $s$ is a slot label, $v$ is the associated value, and $\mS$ is the full set of slots in the ontology.
Values can be associated with a slot-dependent restricted set $\mV_s$ or free-form text (e.g., a home address) or null. 
For multi-domain systems where different domains share some but not all slots (e.g., many domains have a date slot), the domain $d$ is separately tracked:
$\{ (d, s,v), d \in \mD, s \in \mS \}$.
The full set of tuples is updated after each agent-user exchange to support construction of application calls needed to complete the task.

We formalize the our information extraction task as follows. Ignoring domain for brevity, define $(A, U)_t$ as the pair of agent and user turns at exchange $t$. Given a sequence of exchanges between and agent and a user, $\{(A, U)_1, \ldots, (A, U)_t\}$, find the dialogue state $\{ (s,v), s \in \mS_t \}$, where $\mS_t$ is the subset of slots active at time $t$ (i.e., having non-null values). 
The state associated with the final turn $T$ effectively provides a summary of the information extracted from the user in the dialogue.

\subsection{Definition of Extracted Information}
To accommodate the complexities of our dialogues, we augment DST problem in three ways.
First, we introduce the notion of a ``referent'', either with the global context or the entity that the extracted information is associated with. 
Second, we allow slots to take on multiple values. 
Lastly, we allow slot values to be updated in multiple ways: a value can be corrected by the user, a new value can be added to form a list, or an existing value can be augmented, e.g., with details expanding on a free-form slot.
For example, \autoref{fig:teaser} provides an example of an agent gathering information about an accident together with the extracted tuples. There are three referents (\textit{Global} context, \textit{Caller}, and \textit{Other Driver}); the number of passengers in the caller's vehicle was corrected from one to two; and the other driver's car has multiple \textit{Damage Parts} (left and front). 

With these changes, we describe our notations as follows, using the arrow diacritic to indicate cumulative state elements, upper case to indicate tuples and lower case to indicate labels or values, boldface to indicate a set of tuples, and calligraphic font to indicate a set of values.  The initial dialogue state $\bX_0$ is empty.
The cumulative belief (CB) state $\overleftarrow{\bX}_t$ (for $t > 0$) could be predicted directly or via a recursive state update:
$\overleftarrow{\bX}_t = \mathit{update}(\overleftarrow{\bX}_{t-1},\bX_t)$, where only new/updated state values are predicted in 
the turn-level belief (TLB) $\bX_t$ and the update function adds new slots and replaces updated slots.
In the direct approach, it is possible to correct errors made by the model in previous turns, as well as introduce errors.
A potential advantage of the update approach is that TLBs are shorter and therefore easier to predict. 

Formally, $\overleftarrow{\bX}_t$ and $\bX_t$ are defined as follows.  Define $\overleftarrow{\mR}_t$ as the set of referents mentioned in a dialogue up through turn $t$, and $\mR_t \subseteq \overleftarrow{\mR}_t$ as the subset of referents associated with information updates in turn $t$.\footnote{Our application uses a finite set of types $\overleftarrow{\mR}_t \subseteq \mR$, but it could be an open set, e.g., based on names.} 
The dialogue state and TLB after turn $t$, $\overleftarrow{\bX}_t$ and $\bX_t$, respectively, can both be represented as a set of referent-associated sets of active slots:
$$\overleftarrow{\bX}_t = \{ (r, \overleftarrow{\bS}_{rt}), r \in \overleftarrow{\mR}_t\} \ \  \bX_t = \{ (r, \bS_{rt}), r \in \mR_t\}$$
where $\bS_{rt} = \{ S_{r1}, \ldots, S_{r{n_{rt}}}\}$, $n_{rt}$ is the 
number of active slots for referent $r$ updated at turn $t$, and $\overleftarrow{\bS}_{rt}$ denotes the cumulative set of slots.
An active slot is defined as
$S_{rj} = (s_{rj}, \mV_{rj})$, where $s_{rj} \in \mS$ is the $j$th slot linked to referent $r$, $\mS$ is the set of slot (or domain-slot) types, and $\mV_{rj}$ is a set of one or more values $v$ (categorical or free-form text) associated with that slot. 
For our generated data, annotators are asked to provide the state updates.

\input{sections/3.1_evaluation}

%% file: sections/3.1_evaluation.tex
\subsection{Evaluation}
\label{subsec:eval}
In information extraction (IE) tasks, precision, recall, and F-measure are commonly used, while dialogue state tracking (DST) relies on joint goal accuracy (JGA) and slot accuracy.
Similar to DST, our IE task updates extracted information across turns.
Directly adopting DST metrics for dialogue-based IE is not ideal, because they overemphasize earlier parts of a conversation and do not disentangle the effects of error propagation across turns \citep{kim-etal-2022-mismatch}.
For these reasons, we propose to use precision, recall, and $F_1$ scores, along with reporting both cumulative and turn update scores.

Our task requires the scoring to handle multi-value and extended free-form text responses. For scoring purposes, we treat multi-value slots as multiple instances of a slot. For free-form values, following the multi-span setup in question answering~\cite{li-etal-2022-multispanqa}, we enumerate all possible alignments between predicted and gold values. Each gold value is aligned to one predicted value at most, and percentage match is computed based on the longest common substring (LCS) to give a partial-credit score in the range $[0,1]$ (rather than requiring exact match, i.e., $\{ 0,1\}$ score) for use in measuring precision and recall.

\paragraph{Cumulative Score (evaluating $\overleftarrow{\bX}$).}
A cumulative belief (CB) state score $m$ is computed for a particular turn (specific index $t$ or dialogue-final turn) in the $n$th dialogue as follows:
\begin{equation*}
m_{\textsc{cb}}(n,t) = 
    \textstyle\frac{1}{|\overleftarrow{\mR}_{nt}|}\sum_{r\in \overleftarrow{\mR}_{nt}} m(\hat{\overleftarrow{\bS}}_{nrt},\overleftarrow{\bS}^*_{nrt}) .
\end{equation*}
\vspace{-1mm}
where $m$ can be precision ($P$) or recall ($R$). 
Overall scores are obtained by averaging over all dialogues $\mN_t = \{n: \overleftarrow{\mR}_{nt} \ne \emptyset\}$.\footnote{In the first turns, it is possible that there is nothing to extract and no false predictions, in which case $\overleftarrow{\mR}_{nt} = \emptyset$.}
For example, precision is given by:
\vspace{-1mm}
\begin{equation*}
    \textsc{cb-}P(t) = \textstyle\frac{1}{|\mN_t|}\sum_{n\in \mN_t} P_{\textsc{cb}}(n,t).
\end{equation*}
\vspace{-1mm}
We compute the $F_1$ score after getting the averaged precision and recall.

\paragraph{Turn Update Scores (evaluating $\bX$).}  \label{sec:tlb_metrics}
Several scores are computed at the turn level, all of which are based on averaging over all $N$ dialogues in the test set as follows:
\vspace{-1mm}
\begin{equation*}
    \textstyle\frac{1}{N}\sum_n \frac{1}{|\mT_n|}\sum_{t\in \mT_n}  m_{\textsc{type}}(n,t) 
\end{equation*}
where $\mT_n = \{t: \mR_{nt}\ne\emptyset\}$ and $\textsc{type} \in \{\textsc{tlb}, \textsc{r}, \textsc{rs}, \textsc{sv}\}$ denotes diagnostic score type. 
Specific scores ($m_\textsc{type}$) are based on:
\vspace{-1mm}
\begin{align*}
m_{\textsc{tlb}}(n,t) & = \textstyle\frac{1}{|\mR_{nt}|}\sum_{r\in \mR_{nt}}m(\hat{\bS}_{nrt},\bS^*_{nrt})\\
m_{\textsc{r}}(n,t) & = m(\hat{\mR}_{nt}, \mR_{nt}^{*})\\
m_{\textsc{rs}}(n,t) & = \textstyle\frac{1}{|\mR_{nt}|}\sum_{r\in \mR_{nt}}m(\hat{\mS}_{nrt}, \mS^*_{nrt})\\
m_{\textsc{sv}}(n,t) & = m\left(\textstyle \bigcup_{r \in \mR_{nt}}\hat{\bS}_{nrt}, \textstyle\bigcup_{r \in \mR_{nt}}\bS^*_{nrt}\right)
\end{align*}
where $\mS_{nrt}$ is the set of slot labels associated with referent $r$ in turn $t$ of the $n$-th dialogue.
For each turn, the $m_{\textsc{tlb}}$ indicates performance over the TLB; $m_{\textsc{r}}$ indicates how well referents are recognized; $m_{\textsc{rs}}$ indicates how well referents are associated with slots ignoring values; and $m_{\textsc{sv}}$ gives performance of slot-value detection ignoring referents.

%% file: sections/4_dataset.tex
\section{Datasets}
\label{sec:datasets}
\input{tables/datasets_comparison}
We were provided with a private dataset of 34 natural auto insurance claim calls (\aic). In each call, the agent's task is to gather detailed information about an auto accident. The calls were human transcribed and labeled using a schema with 6 referents and 60 possible slots from 10 domains (Appendix \ref{app:as_ontology}). Calls had high variance in length and complexity, as shown in Table~\ref{tab:datasets_comparison}.
Additionally, 50\% of dialogues had multiple values for at least one active slot.
We split the calls into 7/4/23 for train/val./test sets aiming for a slot count split of 20/10/70.

Using \aic as a target dataset for augmentation, we apply \dialgen with ChatGPT
as the LM backbone to create \syntaic, which contains 235 labeled dialogues (Appendix~\ref{app:sample_synt_dials}). 
Reviewers complete a one-hour training to become familiar with the task and practiced generating one dialogue under supervision. 
Full training is complete after they receive feedback for their first 3--5 dialogues. They are instructed to aim for generating dialogues with $\approx$ 50 turns. 
On average, each dialogue comprises 8$\pm$4 subdialogues, with 58\% of edited turns and 20\% of generated turns being deleted. Each dialogue involves $9\pm10$ times of partial or full subdialogue regeneration.

Data collection occurred over 2 months with multiple iterations as documentation and task instructions evolved to become more comprehensive and consistent. The final version of the task instructions further encouraged workers to update slot values in multiple ways and include multiple values in a slot (as described in \S 2.1). We follow the methodology in SQuAD \citep{rajpurkar-etal-2016-squad}, calculating inter-annotator agreement (IAA) at the turn level with three annotators and 32 dialogues, with a resulting IAA of 78.5\% $F_1$ (Appendix~\ref{app:syntaic_iaa_details}). 

\syntaic has less variance than \aic across all statistics, which follows expectations of natural data being noisy and difficult to emulate. However, compared to MultiWOZ \citep{budzianowski2018multiwoz}, \syntaic is more complex. MultiWOZ dialogues average  14 turns and 8 active slots per dialogue, compared to 46 turns and 38 slots on average for {\syntaic}.

We split \syntaic into train/val./test sets with a ratio of 80/10/10 dialogues, selecting val./test sets by randomly sampling from the final iteration of data collection. Table~\ref{tab:datasets_comparison} contains additional statistics of \aic and \syntaic.

%% file: tables/datasets_comparison.tex
\begin{table}[!thp]
\centering
\resizebox{0.95\linewidth}{!}{
\begin{tabular}{l|cc}
\toprule
                              &\aic               &\syntaic \\\midrule
\# dial.                      & 34                & 235 \\
\# turns / dial.              & 197 $\pm$ 98      & 46 $\pm$ 8 \\
\# tokens / dial.             & 4195 $\pm$ 2404   & 1128 $\pm$ 230 \\
\# user tokens / turn         & 18 $\pm$ 27       & 22 $\pm$ 17 \\
\# agent tokens / turn        & 25 $\pm$ 31       & 27 $\pm$ 14 \\
\# referent-slot pair         & 1622              & 8844 \\
\# unique referent-slot       & 109               & 152 \\
\# referent-slot pair / dial. & 48 $\pm$ 24       & 38 $\pm$ 8 \\
\% dial. w/ updates           & 50.0\%            & 14.5\% \\
\% dial. w/ multiple val.     & 50.0\%            & 19.1\% \\
\bottomrule
\end{tabular}
}
\caption{Statistics are calculated on the full dataset. Tokens are calculated with Huggingface T5 tokenizer.
}
\label{tab:datasets_comparison}
\end{table}

%% file: sections/5_experiments.tex
\section{Experiments}
\subsection{Models} 
\paragraph{In-context Learning.}
\citet{hu2022context} propose IC-DST and use schema prompts and a specialized retriever to enable few-shot in-context learning to predict state change with an LM. 
Given longer dialogues, a more complex ontology, and more slots to track than the datasets discussed in \citet{hu2022context}, the representation of dialogue history becomes a crucial concern. 
The SQL tables of the ontology is 1696 tokens, 
and our chosen LM, ChatGPT, has a token limit of 4096 tokens. To accommodate the token constraints, we truncate the in-context examples when given a longer dialogue state. We extract the TLB at turn $t$ and accumulate TLBs as CB.

Furthermore, our task requires the model to identify the corresponding entity (referent) for the predicted slot-value pair.
We redesign the prompt (Appendix~\ref{app:ic_dst_prompt}) to instruct the LM to generate the referent, slot, and value simultaneously.
The retriever, SBERT \cite{reimers-gurevych-2019-sentence}, is finetuned on the full \syntaic training set, which is also used as the example selection pool. Due to privacy concerns, we only evaluate IC-DST on the \syntaic test set.

\paragraph{Finetuned Transformers.}
We follow idea of the previous work \cite{lee2021dialogue} to independently extracted the information and finetune T5 \cite{raffel2020exploring} and Long-T5 \cite{guo-etal-2022-longt5} with schema information embedded in the prompt.
However, unlike the independent decoding in \citet{lee2021dialogue} which used separate prompts for each domain-slot pair, we take a more efficient approach with one prompt per domain, where the model predicts only active slots (together with referent and value). The CB is the aggregate of predictions over all domains.

In addition, we explore four different configurations of prompt and model outputs:
\vspace{-1mm}
\begin{description}
\item[Long-T5\dag:] Use $\{(A,U)_\tau\}_{\tau=1}^{t-1}$ to predict CB 
\vspace{-2mm}
\item[Long-T5:] Use $\{(A,U)_\tau\}_{\tau=1}^{t-1}$ to predict TLB; add to CB 
\vspace{-2mm}
\item[T5:] Use $(A, U)_{t-1}$ to predict TLB; add to CB 
\vspace{-2mm}
\item[T5-SC:] Use $(A, U)_{t-1}$ and previous domain CB to predict state change $\Delta$CB; update CB
\vspace{-1mm}
\end{description}
Due to the input length can be longer than 1k tokens, we choose Long-T5 to cover all turns with the prompt, while the T5-based models make prediction based on the current turn only. T5-SC further considers the state change $\Delta$CB which is similar to the TLB but augmented with the four state-change commands.
Details of prompts for the different cases are given Appendix~\ref{app:sgp_dst_prompt}.

\subsection{Experimental Setup}
When conducting experiments involving \aic, the model selection criterion is the highest TLB $F_1$ score on the \aic validation set. For experiments solely on \syntaic, models were chosen based on TLB $F_1$ score on the \syntaic validation set. Additional hyperparameters can be found in Appendix~\ref{app:sgp-dst_training_details}. All reported values represent the medians of five different random seeds.

%% file: sections/5.1_results.tex
\subsection{Results}
We report results on both cumulative and turn update scores. The cumulative socres are presented in two ways: \Qavg as an average of CB across every user turn, and $CB_Q$ as the CB at user turn $t$, where $t = \ceil{QT/4}, Q\in \{1,2,3,4\}$ and $T$ is the total length of a dialogue. Thus, $t$ will be a specific turn, at either a quarter, a half, three-quarters, or the end of the dialogue.

The score of the last cumulative belief state \Qc is the full $F_1$ score and can be regarded as evaluating a conversation summary.
Model development was done only on the synthetic data to minimize use of real data. 
 
\input{tables/experiments/DG_CB_TLB_v1.8}
\input{tables/experiments/AIC_CB_TLB_v1.8}

\paragraph{Results on \syntaic Test Set.}
The results of experiments on \syntaic with different learning strategies and T5 configurations are presented in \autoref{tab:DG_DST}.
The performance of IC-DST is lower than all T5 variants, although this may be due to the difference in use of domain-specific prompts.
Note that our IC-DST implementation is based on the same ChatGPT model used for generating the \syntaic, so the low results suggest that human collaboration leads to data that is sufficiently different from ChatGPT text such that ChatGPT cannot easily address this task.
Predicting CB directly requires the full history, which is only possible with Long-T5. With Long-T5, there is a benefit to predicting CB directly over TLB. However, optimizations needed to handle a longer history have tradeoffs that result in performance that is worse than the standard T5 model with TLB prediction for this task.
The best result is obtained with T5-SC, which updates values rather than simply adding them as new elements in a list.

To mitigate the potential risk of LMs generating personal information linked to randomly generated names in shared data, we replace them with other randomly generated names.
As shown in \autoref{tab:DG_DST}, T5-SC exhibits comparable performance on both the original and renamed dialogues, indicating that the renaming process does not impact the model's effectiveness.

\input{figure_tex/cb_box_plot}
\input{figure_tex/tlb_percent_curves}
\paragraph{Results on \aic Test Set.}
The two best models (T5 and T5-SC) are used in experiments on the real data (\aic). The $F_1$ results for different training sources are given in \autoref{tab:aic_f1}. 
We measure the utility of synthetic data on model performance by varying amounts of \syntaic. The performance for the model trained on the synthetic data alone is better than with the small amount of the real data, but the best results are obtained by model trained on the combined data. 
Because of the higher frequency of state changes in the human-human dialogues, there is a greater benefit from the T5-SC model for the real data, with an 8\% improvement in the \Qc score compared to 4\% for the synthetic data when using all training data.

To provide more insight into performance, we present the precision/recall results for $\textsc{CB}$ in \autoref{fig:cb_box_plot}.
Incorporating synthetic data yields higher recall and outperforms using real data alone in terms of $F_1$. 
The increased recall can be attributed to the inclusion of a wider range of values in the synthetic data, which are not covered by the \aic training set. However, this improvement comes at the expense of lower precision. By combining both data sets, the model achieves better alignment with real-world data while retaining the advantage of high recall scores from the synthetic data.

We also experimented with varying the amount of synthetic data used in training the model in order to ascertain the relative value of synthetic vs.\ real data. \autoref{fig:tlb_percent_curves} shows that using 59 synthetic dialogues (approximately 2.7K turns) yields results similar to those obtained from the \aic training set, which consists of 1.3K turns in 7 dialogues. These results suggest that roughly 2.1 times as many turns of synthetic data is needed to match the performance of the real data, or 8.4 times as many synthetic dialogues since the synthetic dialogues are shorter. However, the synthetic data is more valuable in combination with real data, for which the benefit beyond 97 dialogues (50\%) is minimal.  This suggests an opportunity for further improvement through strategic scenario sampling. 

%% file: tables/experiments/DG_CB_TLB_v1.8.tex
\begin{table}[!htp]
\centering
\resizebox{0.95\linewidth}{!}{
\begin{tabular}{lcccccc}\toprule
Method &\Qavg &\Qu &\Qd &\Qt &\Qc &TLB \\
\midrule
IC-DST  &71.3 &71.9 &68.5 &68.4 &68.2 &68.1 \\
Long-T5\dag &71.8 &72.5 &71.7 &71.0 &70.4 &-- \\
Long-T5 &66.3 &64.3 &64.8 &64.3 &63.9 &68.5 \\
T5      &76.8 &78.4 &74.9 &73.7 &74.1 &73.9 \\
T5-SC   &\bf78.2 &\bf79.3 &\bf76.4 &\bf76.6 &\bf76.9 &\bf74.2 \\\midrule
T5-SC\S &78.5 &78.7 &76.2 &76.0 &76.2 &75.0 \\
\bottomrule
\end{tabular}
}
\caption{
$F_1$ scores on the \syntaic test set. \dag~denotes Long-T5 with direct CB prediction. \S~denotes the results on the test set with name substitution.
}
\label{tab:DG_DST}
\end{table}

%% file: tables/experiments/AIC_CB_TLB_v1.8.tex
\begin{table}[!htp]
\centering
\resizebox{\linewidth}{!}{
\begin{tabular}{lrrrrrrr}\toprule
Method & Data &\Qavg &\Qu &\Qd &\Qt &\Qc &TLB \\\midrule
T5 & \aic & 38.3 &39.6 &37.1 &36.2 &35.1 &34.8 \\
T5 & DG & 40.4 &41.7 &42.6 &39.9 &37.7 &40.9 \\
T5 & Both & 43.7 &42.9 &42.2 &43.0 &41.9 &43.7 \\\midrule
T5-SC & \aic &39.2 &40.0 &38.1 &37.1 &36.1 &33.9 \\
T5-SC & DG &41.0 &43.6 &42.1 &41.3 &40.5 &38.9 \\
T5-SC & Both &\bf46.2 &\bf47.8 &\bf47.2 &\bf45.9 &\bf45.3 &\bf44.6 \\
\bottomrule
\end{tabular}
}
\caption{
$F_1$ scores on the \aic test set for different training data. DG stands for \syntaic. Both means the data includes \aic and \syntaic.}
\label{tab:aic_f1}
\end{table}

%% file: figure_tex/cb_box_plot.tex
\begin{figure}[ht]
    \centering
    \includegraphics[width=0.85\linewidth]{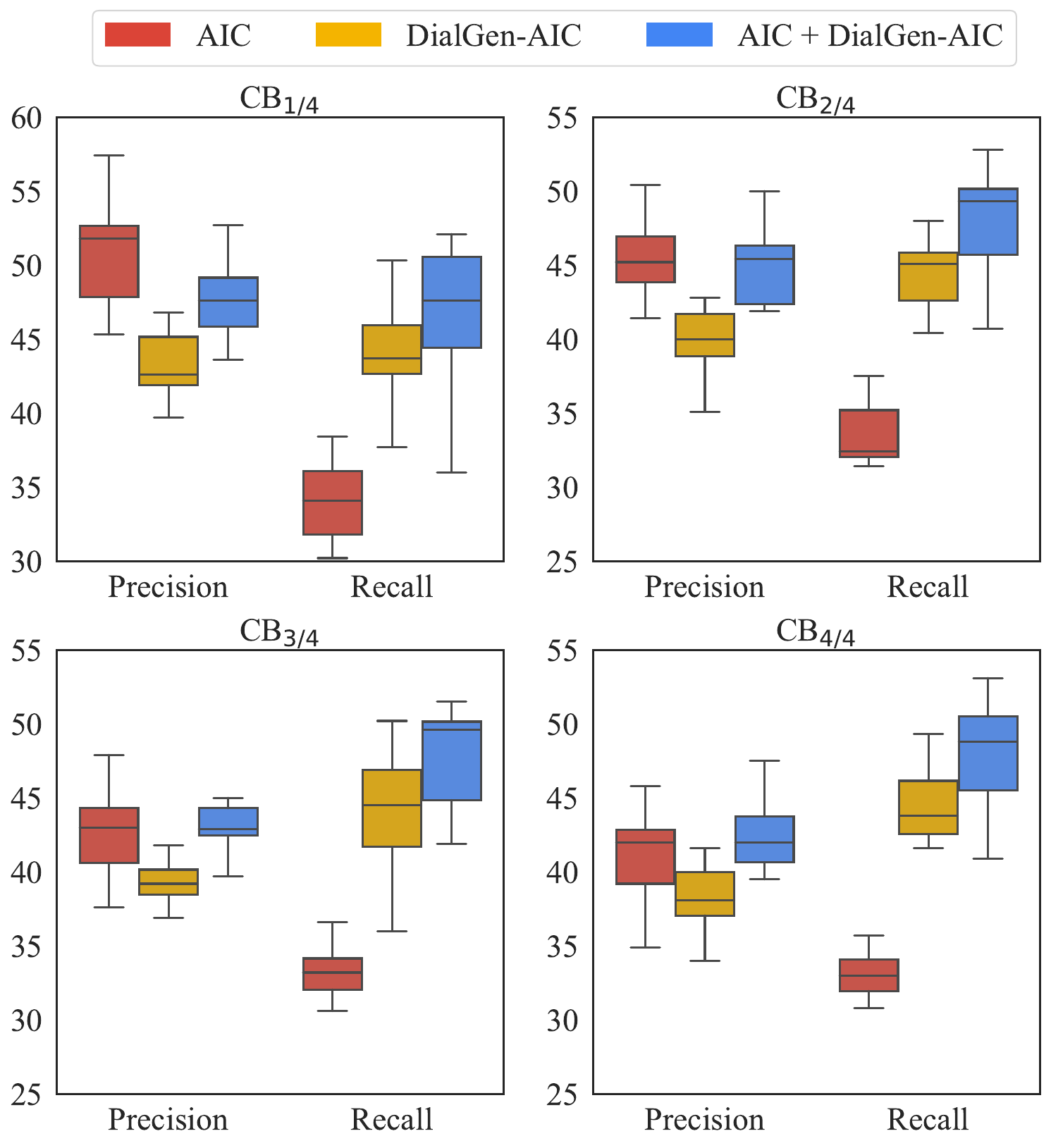}
    \caption{CB precision and recall scores on the \aic test set. All scores are based on T5-SC models.}
    \label{fig:cb_box_plot}
\end{figure}

%% file: figure_tex/tlb_percent_curves.tex
\begin{figure}[ht]
    \centering
    \includegraphics[width=0.95\linewidth]{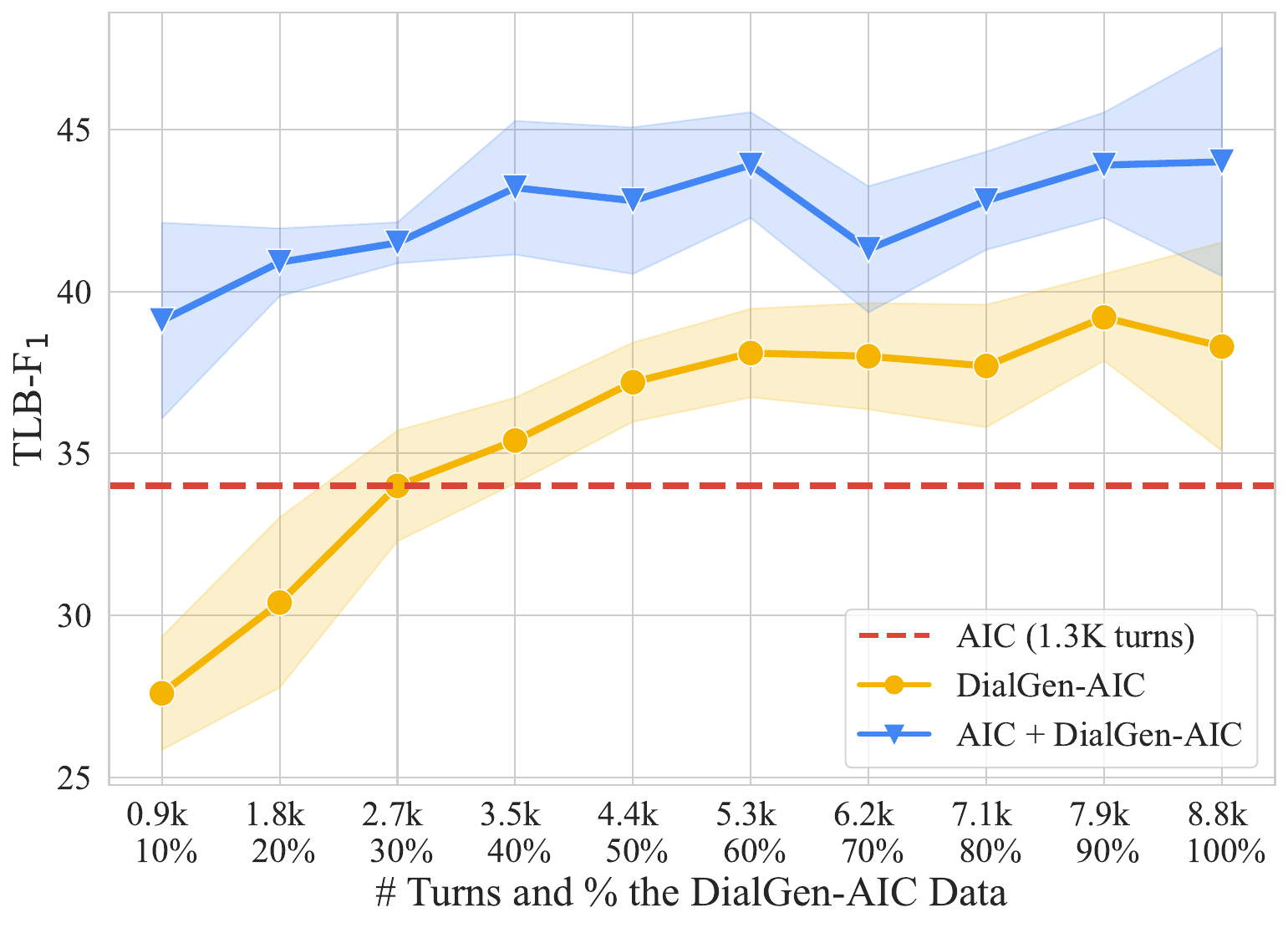}
    \caption{$\textsc{TLB-}F_1$ scores for T5-SC on \aic test set by varying the amount of \syntaic training data.}
    \label{fig:tlb_percent_curves}
\end{figure}

%% file: sections/6_analysis.tex
\section{Error Analysis}
Out of the 56 slots in the AIC test set, we noticed an improvement in 45 slots, while 4 slots were tied, and the remaining 7 slots have slightly worse performance.
Our error analysis reveals two main categories for the performance loss: data mismatch between \aic and \syntaic and over-reliance on surface-level features.

\paragraph{Data Mismatch.} We lose performance for the slot \textit{Car Mileage} because of a difference in language used when describing the mileage of a car. In \aic, agents ask a binary confirmation for whether the mileage on the vehicle is above a certain threshold, whereas callers in \syntaic describe car mileage with an exact number. 
For the slot \textit{Traffic Controls Obeyed}, AIC callers indirectly indicate that traffic controls are not obeyed, e.g. stating that the other driver ran a red light.
In \syntaic, the agent asks the caller to confirm directly whether traffic controls were obeyed. 

\paragraph{Surface Level Text.} The model both over-~and under-predicts slots due to surface-level features such as predicting \textit{Number of Involved Cars} when the text discusses counting vehicles, despite many such instances in \aic simply describing the traffic environment to contextualize the accident, e.g., there was a vehicle in front of the caller, but it was not involved in the accident. The model also predicted this slot when there was language about the number of passengers with a driver. Similarly, \textit{Color} would be predicted whenever colors were mentioned, e.g., a purple bruise. \textit{Traffic Flow} was severely under-predicted when it would have been beneficial for the model to predict the slot whenever it saw information describing lane direction. 

%% file: sections/8_conclusion.tex
\section{Conclusion} 

We propose \dialgen, in which humans and LMs collaborate to generate long, complex dialogues. 
We demonstrate its effectiveness by synthesizing auto insurance calls and conducting information extraction experiments. While we build on the DST framework, our information extraction experiments target an ontology and data that are more complex than the DST task was originally designed for. 
To serve the IE task,
we introduce an entity-centric scoring methodology more suitable for our information extraction task than the conventional joint goal accuracy metrics used in DST. 
Our experiments demonstrate that the data generated by \dialgen, despite dissimilarities with the data it is designed to emulate, can significantly improve model performance for information extraction on real-world human dialogues.

%% file: sections/9_limitations.tex
\section{Limitations}
While \dialgen can be used to generate synthetic data for privacy-constrained settings, the effectiveness largely depends on the LM employed, target setting, and language. We conducted all experiments in the auto insurance claim calls domain in English, where English is a high-resource language, and descriptions of car accidents are reasonably frequent in online text. An LM without reasonable capability in generating text in the target domain and language will result in low quality subdialogues, which can result in a frustrating collaboration for the human reviewer.

Subdialogue generation in \dialgen is guided by including the full dialogue history as context for each subsequent subdialogue. LMs have finite context input length, so the max length of a generated dialogue is limited by the chosen LM. Methods to overcome this limitation can include truncating the dialogue history context, investigating which parts of the prompt contribute little to guiding the LM, and representing dialogue history in a more efficient manner.

\section{Ethical Considerations}
Preserving privacy \citep{flgan,privacyPreserving-rec-sys,diff-private-syn-med-data-cgan} is an important challenge in synthetic data generation. Ensuring important characteristics in synthesized data with \dialgen requires a domain expert who may have access to real, private data and can unintentionally leak information.
\syntaic, on the other hand, generates personal information using the Faker package,\footnote{\url{https://github.com/joke2k/faker}} but there is a potential for the LM to produce personal details related to randomly created names. 
To mitigate the potential risk in shared data, we use gender guesser package \footnote{\url{https://github.com/lead-ratings/gender-guesser}} to detect the gender of each name and replace it with other same-gender name.
If \dialgen users plan to publicly release their data, they should remove potentially identifying information such as names from the synthesized data. 
In the released \syntaic, we replace names with random alternatives to prevent the inadvertent generation of sensitive personal information by the LM. 

Other than privacy issues, LMs can produce harmful content, and the risks of such production can increase depending on the target data setting. When employing humans to collaborate with LMs, practitioners should determine whether additional safety features such as toxic language filters are required to protect the workers.

Regarding the data collection hiring process, all dialogue reviewers were recruited from university listings and compensated at a rate of \$18.69 per hour, following university practices.
Prior to data collection, we instructed our reviewers to familiarize them with the ontology, annotation guidelines, and criteria for assessing dialogue quality.
We established a Slack workspace for smooth communication with the workers throughout the process, providing feedback and promptly addressing questions and concerns they raised.
This interaction ensured high quality of the gathered data.

\section*{Acknowledgments}
We would like to express our sincere gratitude to Kevin Everson, Yanda Chen, and Yushi Hu for their invaluable discussions and preliminary studies. We would also like to thank Bing-Syuan Wang and Irene Wang for their expert web programming consulting and debugging support. Additionally, we extend our appreciation to members of UWNLP for their valuable insights and contributions throughout the project. Lastly, we are grateful to the diligent student reviewers from the University of Washington for their dedicated efforts in data creation. Their contributions were essential to the success of this research.

%% file: sections/appendix.tex
\pagebreak

\section{Training and Generation Details}

\subsection{Finetuning Detains}
\label{app:sgp-dst_training_details}
All experiments are done with T5-base or Long-T5-base with Huggingface implementation \cite{wolf-etal-2020-transformers}. Training time for full \syntaic and \aic setting is averaged 3 hours on 2 NVIDIA V100 GPUs. For the experiments on only \syntaic, we use 2 NVIDIA A40 GPUs.
The total number of GPU training hours is 110 hours.
\input{tables/appendix_utils/t5_training_details}

\subsection{ChatGPT Generation Hyperparameters}
\input{tables/appendix_utils/chatgpt_hyperparameters}

\input{tables/prompts/instruction_prompts}

\section{Prompts}
\label{app:prompt-details}
We shows the prompts used in \dialgen for generating \syntaic, IC-DST, T5 and Long-T5 in the following subsections.

\subsection{\dialgen Prompt}
\label{appendix:dialogue-generation-prompt}
\autoref{tab:personality} shows the list of predefined callers' personality. \autoref{tab:dialgen_prompt} shows an example of a prompt used to generate the first subdialogue when using \syntaic for auto insurance claim calls, including a task description, entity-slot-value triplets, an accident story, caller's and agent's personalities and a initial exchange.
\input{tables/prompts/personality}
\input{tables/prompts/dialgen_prompt}

\subsection{IC-DST Prompt and Output}
\label{app:ic_dst_prompt}
Due to the input length limit, we extract the TLB at turn $t$ and accumulate TLBs as CB. Thus, \texttt{[context]} is regarded as empty.
\input{tables/prompts/icdst_prompt}

\subsection{Prompt and Output for Finetuned Models}
\label{app:sgp_dst_prompt}
The previous study \citep{lee2021dialogue} employs independent decoding with natural language prompts for optimal outcomes. However, this approach necessitates the enumeration of all potential combinations of domain-slot pairs during both training and inference. As the ontology grows larger, the computational burden increases linearly. To address this issue, we propose to group slots with the same domain and train the models to predict all active slots with their values and referents simultaneously. 

\paragraph{Long-T5 for CB prediction.} We present a training example for the ``ContactInfo'' domain with full dialogue history at time $t$.
\input{tables/prompts/sgp_prompt_longt5_cb}

\paragraph{Long-T5 and T5 models for TLB prediction.} We present a training example for the ``ContactInfo'' domain with the most recent two turns $(A, U)_t$ at time $t$.

\input{tables/prompts/sgp_prompt}

In the example, the caller (USER) mentions the first and the last name that are under the domain ContactInfo. The model is require to generate the active slots ``First Name'' and ``Last Name'' with the corresponding values ``Bob'' and ``Lee'', and referent ``Caller.'' 

\paragraph{T5 with State Change (T5-SC).}
\label{app:t5-sc}
For T5-SC, the model need to predict entity-slot-value triplets and edit operations associated with the triplets. The final output of a state at time $t$ will be calculated by applying the edit operations on the associated triplets given the previous state at time $t-1$. We consider four edit operations: \texttt{[new]}, \texttt{[same]}, \texttt{[delete]}, and \texttt{[concat]}. We describe the four edit operations in the following paragraph.

If a triplet has not been observed in the previous state, the model is expected to predict \texttt{[new]}.
Conversely, if the triplet has already been mentioned in the previous state, the model must predict \texttt{[same]}.
The \texttt{[delete]} operation is employed when a triplet mentioned in the previous state should be removed. If the value of a referent-slot is updated, then the model predicts both \texttt{[delete]} for the previous value and \texttt{[new]} for the updated value. 
On the other hand, the \texttt{[concat]} operation is used when the value of a triplet needs refinement, such as combining two values, 7 and AM, into a single value 7 AM. 

Due to the input length limit of the T5 model, we use the most recent $k$ turns to create the previous state and omit the slot descriptions in order to cover more entity-slot-value triplets in the previous state. We get the best results when $k = 18$ for \syntaic and $k = 20$ for \aic. We present a training example for the ``AccidentDetails'' domain as follows.

\input{tables/prompts/sgp_prompt_t5sc}

In the example, the agent (SYSTEM) clarifies the date and time with the caller (USER) because the date and time the caller provides are different from the record in the agent's system. The caller admit the provided time and date are wrong. Thus, time and date need to be updated. The previously provided date ``this Monday'' need to be deleted, so we append an operation \texttt{[delete]} after the value. Similarly, we append the operation after the time ``9:00 am.''

\section{\dialgen}
\label{app:dg_as_generation}

\subsection{Data Collection Cost}
The human reviewers were recruited from university listing. They were compensated at a rate of \$18.69 per hour following our institution’s practices. A dialogue, including reviewing synthesizing and annotation processes, required 45-60 minutes, for a final cost per dialogue of \$14-19. 

\subsection{IAA}
\label{app:syntaic_iaa_details}
We follow the methodology in SQuAD \citep{rajpurkar-etal-2016-squad} for calculating IAA. We select 3 trained workers who participated in data generation as our annotators. They annotated 15\% of \syntaic. The average time to label a dialogue was 18 minutes. For every dialogue, one annotator is randomly assigned as the reference. We calculate max-$F_1$ of every predicted tuple for every turn and average over all turns, then average across all dialogues.

\subsection{\aic Ontology}
\label{app:as_ontology}
\input{tables/ontology}
We show the full ontology in \autoref{tab:as_ontology} including domains, slots, and possible values. Possible referents in the \aic ontology: \textit{Global}, \textit{Caller}, \textit{Other Driver}, \textit{Caller's Passenger}, \textit{Other Driver's Passenger}, and \textit{Witness}. All referents could be associated with every domain/slot, although in practice certain information is almost always associated with a particular referent, e.g., Traffic Conditions (heavy, medium, light) always have a \textit{Global} referent.

\input{figure_tex/tlb_box_plot}

\subsection{User Interface for Data Collection}
\label{appendix:user-interface}
We list two main pages of our interface for dialogue generation. They are editing, and labeling steps.

\begin{figure*}[htp!]
    \centering
    \includegraphics[width=0.8\textwidth]{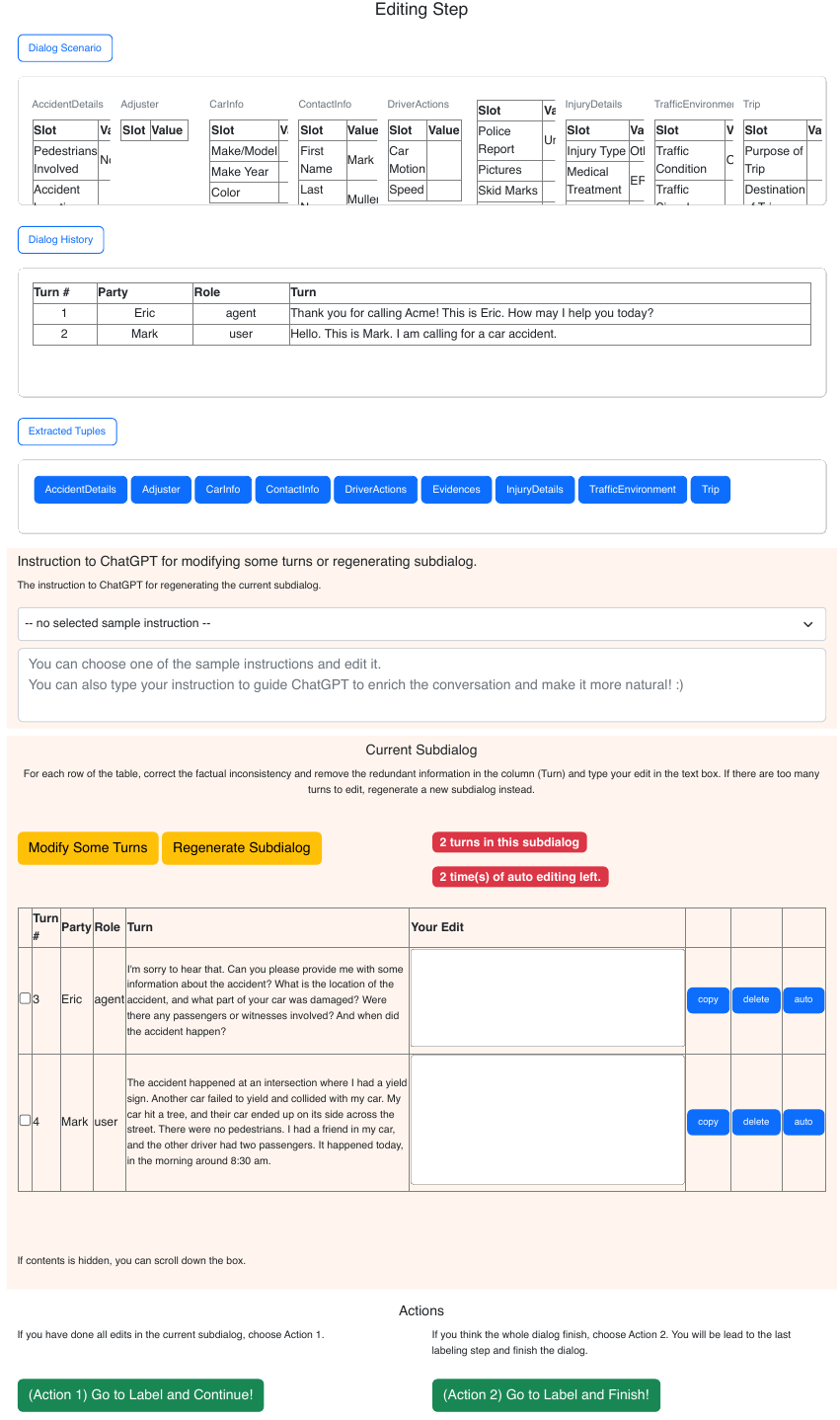}
    \caption{The first step in \dialgen is to create the subdialogue. A dialogue scenario table is provided to indicate slots expected to appear in the conversation. A human reviewer selects LM-generated text and edit it as needed. They can also ask the LM to regenerate selected turns or the full subdialogue and optionally provide extra instructions to guide the LM's generation process.}
    \label{fig:interface_edit}
\end{figure*}

First, the editing step (\autoref{fig:interface_edit}) page provides dialogue scenarios (slot value pairs), dialogue history, extracted tuples (annotated entity-slot-value triplets), instruction for regeneration, and current subdialogue for editing. A human reviewer can provide an instruction to guide the LM to generate a desired subdialogue to replace the current subdialogue. If the the current subdialogue is satisfied with the reviewer, they can edit turns to fix the minor errors in the subdialogue.

\begin{figure*}[htp!]
    \centering
    \includegraphics[width=.95\textwidth]{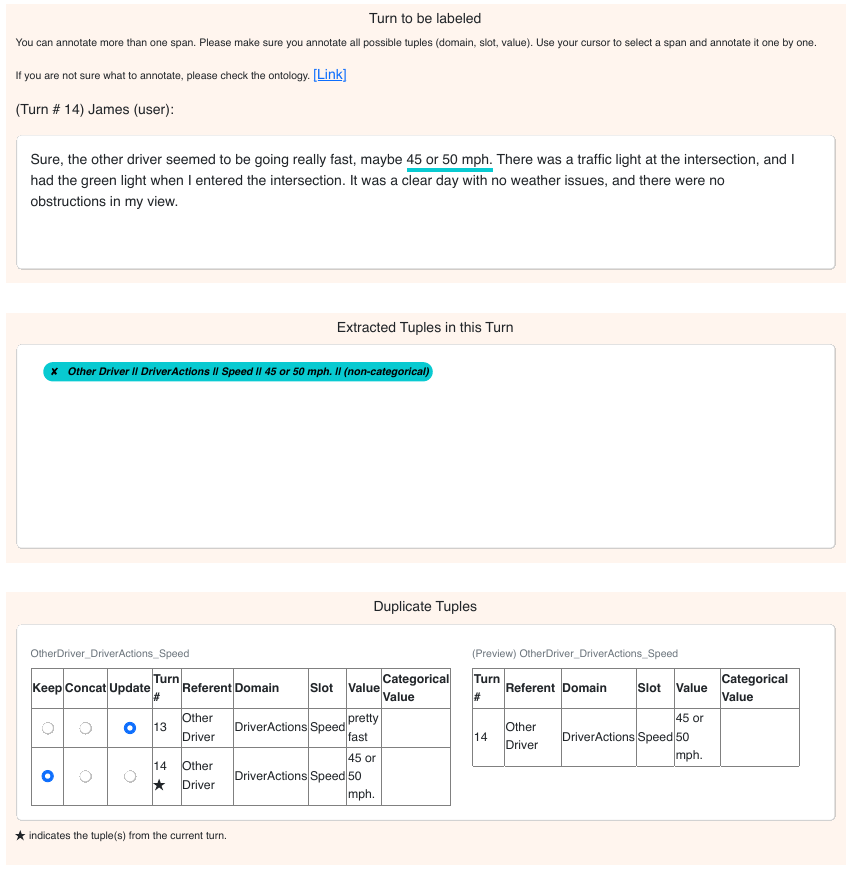}
    \caption{A human reviewer selects a span and label it. If there exists a duplicate label, they are prompted to resolve the conflict by selecting to update (as shown), concat, or keep multiple labels.}
    \label{fig:interface_label}
\end{figure*}

Second, the labeling step page (\autoref{fig:interface_label}) is an optional page for \dialgen framework. This page is designed for dialogue state tracking task where the human reviewer can annotate the edit subdialogue in the previous editing step. Note that the labeling step can be fully decoupled from the framework.

The human reviewer will iteratively collaborate with the LM to generate and revise subdialogues and annotate the subdialogues until reaching the end of the dialogue.

\subsection{\syntaic Dialogues}
\label{app:sample_synt_dials}
In Tables \ref{tab:sample_syntaic_ex1}--\ref{tab:sample_syntaic_ex3}, we show the sample dialogues from \syntaic.
\input{tables/syntaic_examples/ex1}
\input{tables/syntaic_examples/ex2}
\input{tables/syntaic_examples/ex3}

\section{Additional Analysis}
\label{app:pr_analysis}

\autoref{fig:tlb_box_plot} provides the TLB precision and recall results for the full state updates and different diagnostic scores (referent only, referent-slot, and slot-value).  Consistent with the CB results, the biggest benefit of incorporating \syntaic is improved recall. While referent, slot, and value all improve, the greatest improvement is in slot values.

\section{License of Artifacts}
The license of code for \cite{wolf-etal-2020-transformers} is Apache license version 2.0.
The license of code for Faker and Gender-guesser are MIT and GPLv3 License, respectively. The terms for use of our artifacts will be included in our released package.

\clearpage

%% file: tables/appendix_utils/t5_training_details.tex
\begin{table}[H]
\centering
\resizebox{0.85\linewidth}{!}{
\begin{tabular}{l|cc}\toprule
Hyperparameter &T5 &Long-T5 \\\midrule
Training batch size & 16 & 16 \\
Learning rate & $5 \times 10^{-4}$ & $5 \times 10^{-4}$ \\
Max generation length & 256 & 256 \\
Max input length &512 &2592 \\
\bottomrule
\end{tabular}
}
\caption{Hyperparameters for training T5 and Long-T5. The other parameters are default values in Huggingface trainer.}
\label{tab: }
\end{table}

%% file: tables/appendix_utils/chatgpt_hyperparameters.tex
\begin{table}[H]
\centering
\resizebox{\linewidth}{!}{
\begin{tabular}{lcc}
\toprule
Hyperparameter &\dialgen &IC-DST \\\midrule
Version &gpt-3.5-turbo-0301 &gpt-3.5-turbo-0301 \\
Temperature &0.85 - 0.9 &0.0 \\
Max tokens &512 &512 \\
Stop strings & ["<\textbackslash{}div>"] & ["--", "\textbackslash{}n", ";", "\#"] \\
Presence penalty &0.2 &0 \\
Frequency penalty &0.2 &0 \\
\bottomrule
\end{tabular}
}
\caption{Hyperparameters for generation from ChatGPT.}
\label{tab:chatgpt_hyperparameters}
\end{table}

%% file: tables/prompts/instruction_prompts.tex
\begin{table*}[h]
\centering
\resizebox{\linewidth}{!}{
\begin{tabular}{lc}
\toprule
Instruction & Count \\ \midrule
Have CALLER describe more car accident details with complex reasoning that involves two cars' motion. & 23 \\
Have CALLER's response be less specific. have AGENT asks for more details. & 18 \\
Split AGENT's questions into multiple turns & 18 \\
Have CALLER's response be less specific. have AGENT asks for more details. have AGENT asks a question for car accident details. & 15 \\
Have AGENT ask for permission to record the call. & 15 \\
Ask for email address and home address & 14 \\
Have CALLER ask AGENT questions about her insurance coverages in multiple turns & 13 \\
Have AGENT ask CALLER more questions about the accident details & 12 \\
Have CALLER misremember the details. AGENT double check with CALLER. & 12 \\
Explain coverages & 12 \\
Have CALLER corrects wrong information. have AGENT asks for clarification. & 12 \\
Break this conversation down into multiple turns of dialogue & 11 \\
Have AGENT ask for contact information & 10 \\
Break these turns down into multiple turns of back and forth dialogue & 10 \\
AGENT needs to split up her questions. & 10 \\
\bottomrule
\end{tabular}
}
\caption{Instructions with a frequency of 10 or more times used by humans to regenerate a subdialogue.}
\label{tab:instruction_prompts}
\end{table*}

%% file: tables/prompts/personality.tex
\begin{table*}[h]
\centering
\resizebox{\linewidth}{!}{
\begin{tabular}{l|l}
\toprule
\textbf{Personality} & \textbf{Description} \\\midrule
Aggressive & Feeling angry and confrontational about the accident, may place blame on others or use aggressive language. \\
Analytical & Focused on the details and logistics of the claim process, may ask for precise information and explanations. \\
Confused & Unsure about what happened during the accident or what to do next, may ask a lot of questions. \\
Cooperative & Willing to work with the insurance company and other parties involved in resolving the claim. \\
Defensive & Feeling the need to justify their actions or place blame on others, may be unwilling to take responsibility for the accident. \\
Emotional & Experiencing strong emotions related to the accident, may be crying or struggling to maintain composure during the call. \\
Evasive & Hesitant to provide information or answer questions about the accident, may be trying to conceal something. \\
Impatient & Feeling frustrated with the claim process or the speed at which it is progressing, may express irritation or urgency in their language. \\
Reassuring & Trying to maintain a positive and optimistic outlook during the call, may express gratitude for the assistance being provided. \\
Upset & Feeling distressed or frustrated due to the accident and its consequences. \\
\bottomrule
\end{tabular}
}
\caption{The list of the predefined callers' personalities.}
\label{tab:personality}
\end{table*}

%% file: tables/prompts/dialgen_prompt.tex
\begin{table*}[!htp]
\centering
\footnotesize
\resizebox{0.95\linewidth}{!}{
\begin{tabular}{p{\linewidth}}
\toprule
\textless short\_summary\textgreater \newline
story \newline
Bob Parkhurst had a busy day at work, and all he wanted to do was to go grocery shopping. As he backed out of her parking spot in the Office Depot parking lot, he failed to notice the gray MAZDA B-Series Extended Cab driven by Spencer Tullar as he turned into the same aisle from the opposite direction. \newline
Spencer, who was on his way to run some errands, had been driving down the parking lot in extremely slow speed when suddenly he saw Bob's yellow car backing out of his spot. He didn't think much of it and was about to just drive behind her when, at the last minute, he noticed that Bob seemed to be backing out without looking around. Spencer slammed on his brakes, but it was too late. The front right of his truck smashed hard into the back passenger side of Bob's car. \newline
The impact of the collision caused Bob's car to spin around and come to a stop. He immediately felt a sharp pain in her neck and knew that something was wrong. As he tried to get out of the car, he realized that he couldn't move his neck without experiencing excruciating pain. \newline
Spencer got out of his truck and approached Bob's car, he asked if Bob was okay. Bob told him that he was hurt and needed medical attention. Spencer called 911 immediately while also trying his best to comfort Bob until help arrived. \newline
When emergency services arrived shortly after, they found Bob slumped over in her seat, clutching his neck in agony. The responders helped her out of the car and placed a neck brace around him so he wouldn't move his head while they examined her injuries. They then transported him by ambulance to the hospital for further medical attention. \newline
Meanwhile, police were already on their way. Upon arrival at the scene, they took statements from both drivers as well as any witnesses who may have seen what happened. Unfortunately, no one at the time had a clear view of the incident, but both drivers agreed that they didn't see each other before the collision. \newline
Since both cars were still in the parking lot when the accident happened, there was no need to redirect traffic. However, the officers still had to direct people away from the incident site to prevent any further accidents. They also checked Spencer's license and found that it was valid. \newline
The investigation into what caused the accident was inconclusive. Neither driver was certain about who was at fault, as they both believed the other driver failed to observe their movements. Since no one appeared to be at fault, no tickets or \newline
-------- \newline
entity-slot-value triplets \newline
Accident details: (accident location, office depot parking lot), (damage part, unsure), num of passengers, witnesses, date of accident, time of accident, subjective fault, airbag deployed. \newline
Evidences of the car accident: police report, (pictures, no picture), police report number, police department name, tickets citations. \newline
Traffic condition: weather visibility, (obstructions to view, no). \newline
Caller's driver action: car motion, speed, traffic controls obeyed, turn signal, (horn, no). \newline
Caller's car information: (make/model, dodge stratus), make year, color, car mileage. \newline
Caller's injury details: body part injured, injury type, medical treatment. \newline
-------- \newline
task description \newline
Have role play car accident claim call. One person is an agent Alice from a car insurance company and the other is the caller Bob who wants to file a claim. \newline
At beginning of the call, have Alice ask for Bob's permission to record the call and proceeds with the conversation. \newline
Within some \textless p\textgreater \textless /p\textgreater, have simulate poor phone connection. Have Alice and Bob can not hear each other and need to repeat what they said. \newline
Have Alice verify Bob personal information to access account information at the beginning of the call. \newline
Have Bob describe the car accident by using story and tuples above to describe the accident. \newline
Have Alice confirm new information with Bob during the call to ensure consistency. \newline
Have Alice and Bob engage in small talk with each other. \newline
Have Alice explain the insurance coverages to Bob. \newline
-------- \newline
personality \newline
Bob is impatient, feeling frustrated with the claim process or the speed at which it is progressing, may express irritation or urgency in their language. \newline
Alice is conversational, personable, patient, empathetic, sympathetic and professional. \newline
-------- \newline
instructions \newline
Use the story, information, and personality to create a role play script and follow the task description. \newline
\textless /short\_summary\textgreater \newline
\textless div\textgreater \newline
\textless p class="Alice" title="Auto Accident"\textgreater Thank you for calling! This is Alice. How may I help you today? \textless /p\textgreater \newline
\textless p class="Bob" title="Auto Accident"\textgreater Hello. This is Alice. I am calling for a car accident. \textless /p\textgreater \newline
\textless /div\textgreater \newline
Have Alice ask a question for car accident details. \newline
\textless div\textgreater 
\\\bottomrule
\end{tabular}
}
\caption{Example prompt used to generate the first subdialogue in \syntaic. Subsequent subdialogues are generated by appending the previously completed subdialogue to this prompt. Similar to \citet{park2022social}, we use HTML tags to denote different dialogue elements, i.e., <p> for turns and <div> for the subdialogue.}
\label{tab:dialgen_prompt}
\end{table*}

%% file: tables/prompts/icdst_prompt.tex
\begin{lstlisting}[basicstyle=\scriptsize\ttfamily]
CREATE TABLE AccidentDetails(
    'Damage Part' TEXT CHECK ('Damage Part' IN 'Front', 'Right', 'Back', 'Left', 'Front Right', 'Front Left', 'Back Left', 'Back Right', 'Other', 'Unsure'),
    'Accident Location' TEXT CHECK ('Accident Location' IN 'Parking Lot', 'Driveway', 'Highway', 'Roadway', 'Intersection', 'Other'),
    'Num of Passengers' TEXT CHECK ('Num of Passengers' IN '0', '1', '2+', 'Unsure'),
    'Witnesses' TEXT CHECK ('Witnesses' IN 'Yes', 'No', 'Unsure'),
    'Num of Involved Cars' TEXT CHECK ('Num of Involved Cars' IN '1', '2', '3', '4+', 'Unsure'),
    'Children Involved' TEXT CHECK ('Children Involved' IN 'Yes', 'No', 'Unsure'),
    'Airbag Deployed' TEXT CHECK ('Airbag Deployed' IN 'Yes', 'No', 'Unsure'),
    'Towed' TEXT CHECK ('Towed' IN 'Yes', 'No', 'Unsure'),
    'Pedestrians Involved' TEXT CHECK ('Pedestrians Involved' IN 'Yes', 'No', 'Unsure'),
    'Date of Accident' TEXT,
    'Time of Accident' TEXT,
    'Subjective Fault' TEXT CHECK ('Subjective Fault' IN 'Caller', 'Other Driver'),
)

CREATE TABLE Adjuster(
    'Explain Coverages' TEXT,
    'Permission to Record' TEXT CHECK ('Permission to Record' IN 'Yes', 'No'),
    'Set up Inspection' TEXT CHECK ('Set up Inspection' IN 'Quick Photo Claim', 'Field Assignment'),
    'Set up Rental' TEXT CHECK ('Set up Rental' IN 'Yes', 'No'),
)

CREATE TABLE CarInfo(
    'Make/Model' TEXT,
    'Make Year' TEXT,
    'Color' TEXT,
    'Car Mileage' TEXT,
    'Rideshare (Uber/Lyft)' TEXT CHECK ('Rideshare (Uber/Lyft)' IN 'Yes', 'No', 'Unsure'),
)

CREATE TABLE ContactInfo(
    'First Name' TEXT,
    'Last Name' TEXT,
    'Home Address' TEXT,
    'Phone Number' TEXT,
    'Email Address' TEXT,
    'Policy Number' TEXT,
    'Date of Birth' TEXT,
)

CREATE TABLE DriverActions(
    'Car Motion' TEXT CHECK ('Car Motion' IN 'Traveling Forward', 'Backing', 'Turning', 'Changing Lanes', 'Stopped', 'Other', 'Unsure'),
    'Speed' TEXT,
    'Distractions' TEXT CHECK ('Distractions' IN 'Cellphone', 'Animals', 'Smoking', 'Passengers', 'Traffic', 'Eating', 'Not Paying Attention', 'Other', 'Unsure', 'No Distraction'),
    'Brake' TEXT CHECK ('Brake' IN 'Yes', 'No', 'Unsure'),
    'Horn' TEXT CHECK ('Horn' IN 'Yes', 'No', 'Unsure'),
    'Turn Signal' TEXT CHECK ('Turn Signal' IN 'Yes', 'No', 'Unsure'),
    'Traffic Controls Obeyed' TEXT CHECK ('Traffic Controls Obeyed' IN 'Yes', 'No', 'Unsure'),
)

CREATE TABLE Evidences(
    'Police Report' TEXT CHECK ('Police Report' IN 'Yes', 'No', 'Unsure'),
    'Police Department Name' TEXT,
    'Pictures' TEXT CHECK ('Pictures' IN 'At Scene', 'After Accident', 'No Picture', 'Unsure'),
    'Tickets Citations' TEXT CHECK ('Tickets Citations' IN 'Caller Party Cited', 'Other Party Cited', 'No Party Cited', 'Multiple Parties Cited', 'Unsure', 'No Ticket'),
    'Police Report Number' TEXT,
    'Skid Marks' TEXT CHECK ('Skid Marks' IN 'Yes', 'No', 'Unsure'),
)

CREATE TABLE InjuryDetails(
    'Ambulance' TEXT CHECK ('Ambulance' IN 'Yes', 'No', 'Unsure'),
    'Body Part Injured' TEXT CHECK ('Body Part Injured' IN 'Head', 'Neck', 'Shoulder', 'Chest', 'Abdomen', 'Back', 'Limb', 'Other'),
    'Injury Type' TEXT CHECK ('Injury Type' IN 'Bruise', 'Broken Fracture', 'Cut Scratch', 'Bleeding', 'Strain Sprain', 'Sore', 'Other', 'No Injury'),
    'Medical Treatment' TEXT CHECK ('Medical Treatment' IN 'MRI', 'Surgery', 'Cat Scan', 'Hospitalization', 'ER', 'X-Ray', 'Other'),
)

CREATE TABLE TrafficEnvironment(
    'Weather Visibility' TEXT CHECK ('Weather Visibility' IN 'Clear', 'Cloudy', 'Rainy', 'Snowy', 'Foggy', 'Windy', 'Other', 'Unsure'),
    'Obstructions to View' TEXT CHECK ('Obstructions to View' IN 'Yes', 'No', 'Unsure'),
    'Road Condition' TEXT CHECK ('Road Condition' IN 'Dry', 'Wet', 'Slippery', 'Debris', 'Potholes', 'Straight', 'Curved', 'Tunnel', 'Steep Incline', 'Flat', 'Other', 'Unsure'),
    'Traffic Signal' TEXT CHECK ('Traffic Signal' IN 'Stop Sign', 'Yield Sign', 'Green Light', 'Yellow Light', 'Red Light', 'Other', 'Unsure', 'No Signal Or Sign'),
    'Description of Lanes' TEXT CHECK ('Description of Lanes' IN 'Normal', 'Turn Lane', 'Shoulder', 'Other', 'Unsure'),
    'Num of Lanes' TEXT CHECK ('Num of Lanes' IN '1', '2', '3', '4+', 'Unsure'),
    'Traffic Condition' TEXT CHECK ('Traffic Condition' IN 'Heavy', 'Moderate', 'Light', 'Other', 'Unsure'),
    'Speed Limit' TEXT,
    'Traffic Flow' TEXT CHECK ('Traffic Flow' IN 'One-Way', 'Two-Way', 'Other', 'Unsure'),
    'Parking Lot Type' TEXT CHECK ('Parking Lot Type' IN 'Angled', 'Straight', 'Other', 'Unsure'),
)

CREATE TABLE Trip(
    'Destination of Trip' TEXT,
    'Purpose of Trip' TEXT,
    'Origin of Trip' TEXT,
)

-- Using valid SQLite, answer the following multi-turn conversational questions for the tables provided above.

Example #1
[context]
[system] I see. Thank you for letting me know. Can you also provide me with the make, model, and year of your car, as well as its color?
Q: [user] Of course. It's a white Lexus sedan, 2018 model.
SQL: SELECT * FROM CarInfo WHERE Caller-Make_Year = 2018 AND Caller-Color = white AND Caller-Make/Model = Lexus sedan,;

Example #2
[context]
[system] Thank you for sharing that information, Lynne. Can you also provide me with the make and model of your car?
Q: [user] Yes, it's a white sedan. The make and model is a Toyota Camry. It's a 2018 model, and it had about 40,000 miles on it at the time of the accident
.
SQL: SELECT * FROM CarInfo WHERE Caller-Color = white sedan. AND Caller-Make/Model = Toyota Camry. AND Caller-Make_Year = 2018 AND Caller-Car_Mileage = 40,
000;

Example #3
[context]
[system] I see. Can you describe your car's make and model? What year was it made? And what color was it?
Q: [user] It's a white sedan, a 2018 Honda Accord.
SQL: SELECT * FROM CarInfo WHERE Caller-Make/Model = sedan, a 2018 Honda Accord. AND Caller-Make_Year = 2018 AND Caller-Color = white;

Example #4
[context]
[system] Do you remember the make and model of the other car?
Q: [user] I think it was a black sedan, but I'm not completely sure.
SQL: SELECT * FROM CarInfo WHERE Other_Driver-Make/Model = sedan, AND Other_Driver-Color = black;

Example #5
[context]
[system] Thank you for that information, Joel. Can you please provide me with your car's make and model, year, color, and approximate mileage?
Q: [user] Sure, my car is a white sedan. It's a 2016 model with approximately 50,000 miles on it.
SQL: SELECT * FROM CarInfo WHERE Caller-Make/Model = sedan. AND Caller-Car_Mileage = approximately 50,000 miles AND Caller-Color = white AND Caller-Make_Ye
ar = 2016 model;

Example #6
[context]
[system] Thank you for all the details, Richard. Can you please provide me with your car's make and model?
Q: [user] Yes, it's a white sedan, a 2007 make.
SQL: SELECT * FROM
CarInfo WHERE Caller-Color = white sedan AND Caller-Make_Year = 2007
 * FROM CarInfo WHERE Caller-Color = white sedan AND Caller-Make_Year = 2007
 * FROM CarInfo WHERE Caller-Color = white sedan AND Caller-Make_Year = 2007
\end{lstlisting}

%% file: tables/prompts/sgp_prompt_longt5_cb.tex
\label{app:sgp_prompt}
\begin{lstlisting}[basicstyle=\scriptsize\ttfamily]
Input: 
[USER] My name is Bob Lee, and my policy number is 123456789. [SYSTEM] Thank you. Could you please provide me with your name and policy number so I can access your account information? [USER] Yes, that's fine. [SYSTEM] I am so sorry that happened. Before we begin, may I please have your permission to record this call for quality and training purposes? [USER] Hello. This is Bob. I am calling for a car accident. [SYSTEM] Thank you for calling AllState! This is Alice. How may I help you today? [domain] ContactInfo [possible slots] First Name (the First Name of the ContactInfo) [s] Last Name (the Last Name of the ContactInfo) [s] Home Address (the Home Address of the ContactInfo) [s] Phone Number (the Phone Number of the ContactInfo) [s] Email Address (the Email Address of the ContactInfo) [s] Policy Number (the Policy Number of the ContactInfo) [s] Date of Birth (the Date of Birth of the ContactInfo)

Output:
First Name [srv] Bob [rv] Caller [s] Last Name [srv] Lee [rv] Caller [s] Policy Number [srv] 123456789. [rv] Caller
\end{lstlisting}

%% file: tables/prompts/sgp_prompt.tex
\label{app:sgp_prompt}
\begin{lstlisting}[basicstyle=\scriptsize\ttfamily]
Input: 
[USER] Hi, my name is Bob Lee. I was recently in a car accident and wanted to file a claim. [SYSTEM] Thank you for calling! This is Alice. How may I help you today? [domain] ContactInfo [possible slots] First Name (the First Name of the ContactInfo) [s] Last Name (the Last Name of the ContactInfo) [s] Home Address (the Home Address of the ContactInfo) [s] Phone Number (the Phone Number of the ContactInfo) [s] Email Address (the Email Address of the ContactInfo) [s] Policy Number (the Policy Number of the ContactInfo) [s] Date of Birth (the Date of Birth of the ContactInfo)

Output:
First Name [srv] Bob [rv] Caller [s] Last Name [srv] Lee [rv] Caller
\end{lstlisting}

%% file: tables/prompts/sgp_prompt_t5sc.tex
\label{app:sgp_prompt_t5sc}
\begin{lstlisting}[basicstyle=\scriptsize\ttfamily]
Input: 
[USER] Oh, sorry about that. You're right, it actually occurred on a Wednesday at 11 am. [SYSTEM] Also, I just wanted to clarify some information. In our previous conversation, you stated that the accident occurred on a Monday at 9 am. However, our records show that it actually occurred on a Wednesday at 11 am. Can you confirm which day and time the accident actually occurred? [state] Damage Part [srv] Front Left [rv] Caller [cv] Right [rv] Global [s] Accident Location [srv] Highway [rv] Global [s] Num of Passengers [srv] 0 [rv] Global [s] Witnesses [srv] Yes [rv] Global [s] Date of Accident [srv] this Monday [rv] Global [s] Time of Accident [srv] 9:00 am. [rv] Global [s] Subjective Fault [srv] Caller [rv] Caller [domain] AccidentDetails [possible slots] Damage Part [s] Accident Location [s] Num of Passengers [s] Witnesses [s] Num of Involved Cars [s] Children Involved [s] Airbag Deployed [s] Towed [s] Pedestrians Involved [s] Date of Accident [s] Time of Accident [s] Subjective Fault

Output:
Date of Accident [srv] Wednesday [v] this Monday [vo] [delete] [rv] Global [s] Time of Accident [srv] 11 am. [v] 9:00 am. [vo] [delete] [rv] Global
\end{lstlisting}

%% file: tables/ontology.tex
\begin{table*}[!htp]
\centering
\small
\resizebox{\linewidth}{!}{
  \begin{tabular}{lll}\toprule
  Domain &Slot &Possible Values \\\midrule
  Adjuster &Explain Coverages &[] \\
  Adjuster &Permission to Record &[yes, no] \\
  Adjuster &Set up Inspection &[photo claim, field assignment] \\
  Adjuster &Set up Rental &[yes, no] \\ \hline
  ContactInfo &First Name &[] \\
  ContactInfo &Last Name &[] \\
  ContactInfo &Home Address &[] \\
  ContactInfo &Phone Number &[] \\
  ContactInfo &Email Address &[] \\
  ContactInfo &Policy Number &[] \\
  ContactInfo &Date of Birth &[] \\ \hline
  DriverActions &Car Motion &[traveling forward, backing, turning, changing lanes, stopped, other, unsure] \\
  DriverActions &Speed &[] \\
  DriverActions &Distractions &[cellphone, animals, smoking, passengers, traffic, eating, not paying attention, other, unsure, no distraction] \\
  DriverActions &Brake &[yes, no, unsure] \\
  DriverActions &Horn &[yes, no, unsure] \\
  DriverActions &Turn Signal &[yes, no, unsure] \\
  DriverActions &Traffic Controls Obeyed &[yes, no, unsure] \\ \hline
  Evidences &Police Report &[yes, no, unsure] \\
  Evidences &Police Department Name &[] \\
  Evidences &Pictures &[at scene, after accident, no picture, unsure] \\
  Evidences &Tickets Citations &[caller party cited, other party cited, no party cited, multiple parties cited, unsure, no ticket] \\
  Evidences &Police Report Number &[] \\
  Evidences &Skid Marks &[yes, no, unsure] \\ \hline
  InjuryDetails &Ambulance &[yes, no, unsure] \\
  InjuryDetails &Body Part Injured &[head, neck, shoulder, chest, abdomen, back, limb, other] \\
  InjuryDetails &Injury Type &[bruise, broken fracture, cut scratch, bleeding, strain sprain, sore, other, no injury] \\ 
  InjuryDetails &Medical Treatment &[MRI, surgery, CAT scan, hospitalization, ER, x-ray, other] \\ \hline
  AccidentDetails &Damage Part &[front, right, back, left, front right, front left, back left, back right, other, unsure] \\
  AccidentDetails &Accident Location &[parking lot, driveway, highway, roadway, intersection, other] \\
  AccidentDetails &Num of Passengers &[0, 1, 2+, unsure] \\
  AccidentDetails &Witnesses &[yes, no, unsure] \\
  AccidentDetails &Num of Involved Cars &[1, 2, 3, 4+, unsure] \\
  AccidentDetails &Children Involved &[yes, no, unsure] \\
  AccidentDetails &Airbag Deployed &[yes, no, unsure] \\
  AccidentDetails &Towed &[yes, no, unsure] \\
  AccidentDetails &Pedestrians Involved &[yes, no, unsure] \\
  AccidentDetails &Date of Accident &[] \\
  AccidentDetails &Time of Accident &[] \\
  AccidentDetails &Subjective Fault &[caller, other driver] \\ \hline
  CarInfo &Make/Model &[] \\
  CarInfo &Make Year &[] \\
  CarInfo &Color &[] \\
  CarInfo &Car Mileage &[] \\
  CarInfo &Rideshare (Uber/Lyft) &[yes, no, unsure] \\ \hline
  Trip &Destination of Trip &[] \\
  Trip &Purpose of Trip &[] \\
  Trip &Origin of Trip &[] \\ \hline
  TrafficEnvironment &Weather Visibility &[clear, cloudy, rainy, snowy, foggy, windy, other, unsure] \\
  TrafficEnvironment &Obstructions to View &[yes, no, unsure] \\
  TrafficEnvironment &Road Condition &[dry, wet, slippery, debris, potholes, straight, curved, tunnel, steep incline, flat, other, unsure] \\
  TrafficEnvironment &Traffic Signal &[stop sign, yield sign, green light, yellow light, red light, other, unsure, no signal or sign] \\
  TrafficEnvironment &Description of Lanes &[normal, turn lane, shoulder, other, unsure] \\
  TrafficEnvironment &Num of Lanes &[1, 2, 3, 4+, unsure] \\
  TrafficEnvironment &Traffic Condition &[heavy, moderate, light, other, unsure] \\
  TrafficEnvironment &Speed Limit &[] \\
  TrafficEnvironment &Traffic Flow &[one-way, two-way, other, unsure] \\
  TrafficEnvironment &Parking Lot Type &[angled, straight, other, unsure] \\
  \bottomrule
  \end{tabular}
}
\caption{\aic ontology. Empty lists indicate free-form extractive values.}
\label{tab:as_ontology}
\end{table*}

%% file: figure_tex/tlb_box_plot.tex
\begin{figure}[ht]
    \centering
    \includegraphics[width=0.9\linewidth]{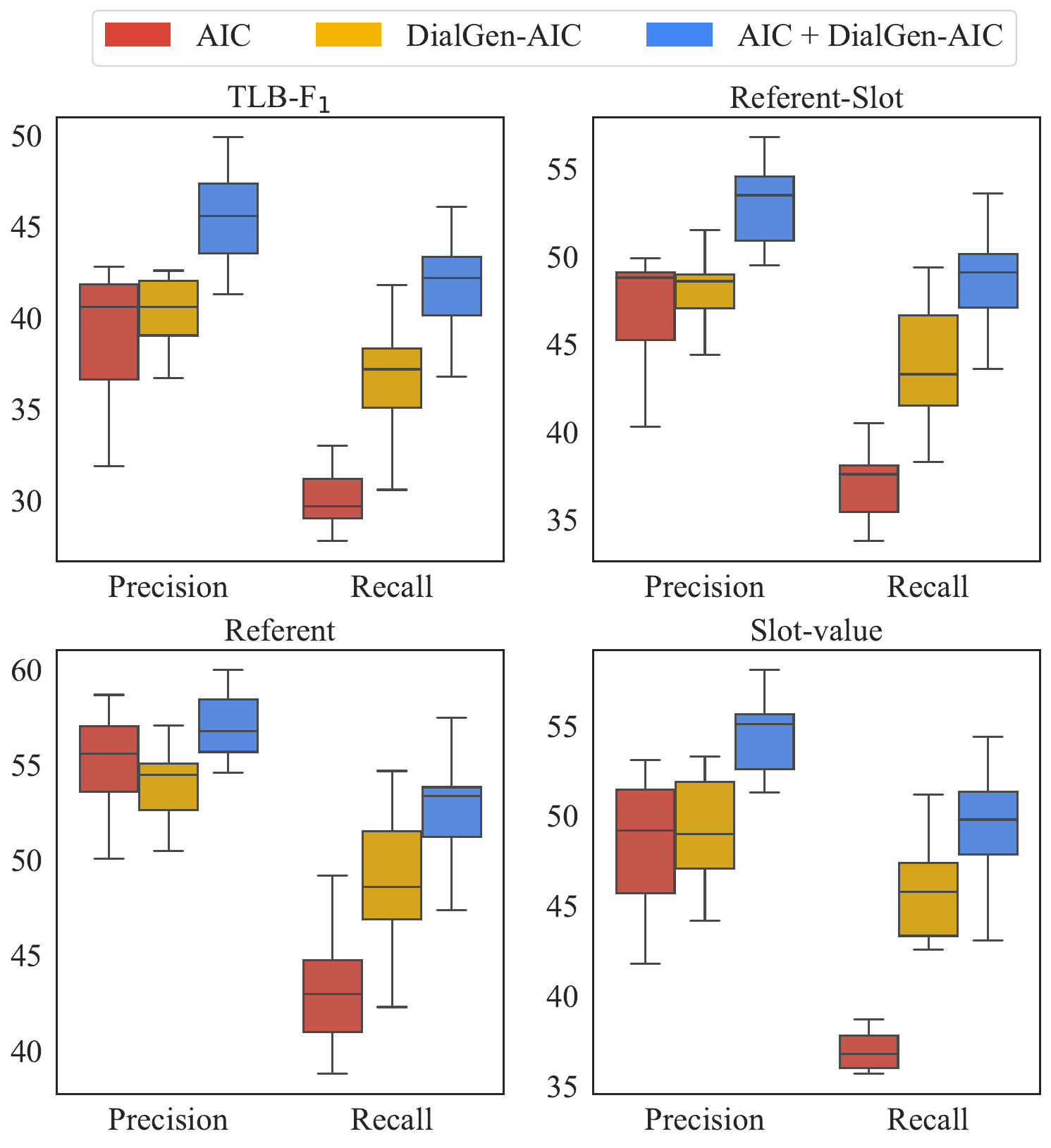}
    \caption{$\textsc{tlb}$ and three diagnostic scores for precision and recall ($m_{\textsc{r}}$, $m_{\textsc{rs}}$, and $m_{\textsc{sv}}$) for the T5-SC model on \aic test set.}
    \label{fig:tlb_box_plot}
\end{figure}

%% file: tables/syntaic_examples/ex1.tex
\begin{table*}[ht]
    \centering
    \footnotesize
    \resizebox{0.98\linewidth}{!}{
    \begin{tabular}{rp{0.95\linewidth}}
    \toprule
    \textcolor{red}{Agent} &Thank you for calling DialGen Insurance! This is Greta. How may I help you today? \\
    \textcolor{blue}{User} &Hello. This is Suzanne. I am calling for a car accident. \\
    \textcolor{red}{Agent} &Alright, Suzanne. First, I need to verify some information. May I please have your full name and policy number? \\
    \textcolor{blue}{User} &Yes, my name is Suzanne Luna and my policy number is 1234567890. \\
    \textcolor{red}{Agent} &Thank you, Suzanne. And can you please provide me with your phone number and address? \\
    \textcolor{blue}{User} &Yes, of course. My phone number is (123) 456-7890, and my address is 123 Main Street. \\
    \textcolor{red}{Agent} &Thank you, Suzanne. Now, could you please describe the accident for me? \\
    \textcolor{blue}{User} &Certainly. I was driving alone on my way to the library when I approached an intersection with a yield sign. I slowed down and checked for oncoming traffic but didn't see any cars, so I began to turn left. Suddenly, I heard a loud crunching sound, and my car jerked to a stop. I got out of my car to assess the damage and saw that a black Nissan Maxima had hit the back of my car. \\
    \textcolor{red}{Agent} &Thank you for letting me know, Suzanne. Do you know the name of the street you were on when the accident happened? \\
    \textcolor{blue}{User} &Sure, the accident happened at the intersection of Main Street and Broad Avenue. \\
    \textcolor{red}{Agent} &And I believe you said you were driving alone, is that correct? \\
    \textcolor{blue}{User} &Yes, I was alone in my car. \\
    \textcolor{red}{Agent} &Were there any witnesses at the scene who provided their contact information? \\
    \textcolor{blue}{User} &Yes, there were several witnesses at the scene who provided their contact information, but I didn't get their names. \\
    \textcolor{red}{Agent} &Do you remember the date and time of the accident? \\
    \textcolor{blue}{User} &Yes, the accident occurred a few nights ago at around 11 pm. \\
    \textcolor{red}{Agent} &Do you know the specific day? \\
    \textcolor{blue}{User} &Let's see, it would have been Wednesday--no, Thursday night. \\
    \textcolor{red}{Agent} &Thank you, Suzanne. Can you tell me if your airbags deployed during the accident? \\
    \textcolor{blue}{User} &No, the airbags did not deploy. \\
    \textcolor{red}{Agent} &Alright, and can you describe the traffic conditions at the time of the accident? Was there a traffic light at the intersection? \\
    \textcolor{blue}{User} &Yes, there was a traffic light and a yield sign. \\
    \textcolor{red}{Agent} &I see. Did you notice anything else about the road or lanes that might have contributed to the accident? \\
    \textcolor{blue}{User} &No, nothing else that I recall. It all happened so fast. \\
    \textcolor{red}{Agent} &Do you remember if you were going at a particular speed when the accident happened? \\
    \textcolor{blue}{User} &No, I was not speeding. I don't think the other driver was either. \\
    \textcolor{red}{Agent} &Alright. Can you tell me about any damage your car sustained? \\
    \textcolor{blue}{User} &Yes, my car sustained significant damage on the left side. \\
    \textcolor{red}{Agent} &I'm sorry to hear that. Do you know if a police report was filed? \\
    \textcolor{blue}{User} &Yes, the Houston Police Department took our statements, and the report number is 1234567890. \\
    \textcolor{red}{Agent} &Do you know if either of you received a citation or ticket as a result of the accident? \\
    \textcolor{blue}{User} &Yes, the other driver, Homer Shepherd, received a citation. \\
    \textcolor{red}{Agent} &Alright, thank you for letting me know. And can you provide me with the make and model of your car? \\
    \textcolor{blue}{User} &Yes, my car is a brown sedan, a Toyota Corolla. \\
    \textcolor{red}{Agent} &Thank you, Suzanne. And can I ask if you sustained any injuries as a result of the accident? \\
    \textcolor{blue}{User} &Yes, unfortunately, I did sustain some injuries. I needed a CT scan at the hospital after the accident. \\
    \textcolor{red}{Agent} &Can you tell me where you were injured, Suzanne? \\
    \textcolor{blue}{User} &My neck and back. \\
    \textcolor{red}{Agent} &What was the diagnosis at the hospital? \\
    \textcolor{blue}{User} &The diagnosis was muscle strain and some minor bruising. \\
    \textcolor{red}{Agent} &I'm sorry to hear that, Suzanne. How are you feeling now? \\
    \textcolor{blue}{User} &I'm doing a bit better, thank you for asking. \\
    \textcolor{red}{Agent} &That's good to hear. Now, let me explain your coverage options to you. \\
    \textcolor{blue}{User} &Okay, thank you. \\
    \textcolor{red}{Agent} &First, let me go over the details of your policy with you. You have liability insurance, which covers bodily injuries and property damage to others if you are at fault in an accident, although it looks like that's not applicable here. You also have collision coverage, which pays for damages to your vehicle in the event of an accident. \\
    \textcolor{blue}{User} &Okay, got it. \\
    \textcolor{red}{Agent} &Additionally, you have personal injury protection, which will cover your medical expenses related to injuries sustained in the accident, regardless of who was at fault. However, please note that there may be limits on the amount of coverage provided by this policy. \\
    \textcolor{blue}{User} &Okay, thank you for explaining that to me. \\
    \textcolor{red}{Agent} &Suzanne, can you tell me if your car was towed from the scene of the accident? \\
    \textcolor{blue}{User} &No, it wasn't. I was able to drive it home myself. \\
    \textcolor{red}{Agent} &Okay, thank you for letting me know. I will just note that in the report. That should be everything for the preliminary details. Is there anything else you'd like to add, Suzanne? \\
    \textcolor{blue}{User} &No, that's all. Thank you for your help, Greta. \\
    \textcolor{red}{Agent} &You're welcome. We will process your claim and be in touch with you soon about next steps. \\
    \textcolor{blue}{User} &Okay, thank you so much. \\
    \textcolor{red}{Agent} &Have a good day, Suzanne. \\
    \textcolor{blue}{User} &You too, Greta. Bye. \\
    \bottomrule
    \end{tabular}
    }
    \caption{Sample \syntaic dialogue 1.}
    \label{tab:sample_syntaic_ex1}
\end{table*}

%% file: tables/syntaic_examples/ex2.tex
\begin{table*}[ht]
    \centering
    \footnotesize
    \resizebox{0.98\linewidth}{!}{
    \begin{tabular}{rp{0.95\linewidth}}
    \toprule
    \textcolor{red}{Agent} &Thank you for calling DialGen Insurance! This is Loan. How may I help you today? \\
    \textcolor{blue}{User} &Hello. This is Joshua. I am calling for a car accident. \\
    \textcolor{red}{Agent} &I'm sorry to hear that you were in an accident. What happened? \\
    \textcolor{blue}{User} &It was last Monday morning in a tunnel. There were several cars involved, and my car was hit from the back by a yellow Dodge Ram. I was injured and had to be taken to the hospital by ambulance. \\
    \textcolor{red}{Agent} &Oh my, I'm sorry to hear that. Did you have any passengers in your car? \\
    \textcolor{blue}{User} &No, I was the only one in the car. \\
    \textcolor{red}{Agent} &Do you know how many cars were involved total? \\
    \textcolor{blue}{User} &I think there were about four cars involved. \\
    \textcolor{red}{Agent} &Okay, thank you for that information, Joshua. Can you describe the damage to your car? \\
    \textcolor{blue}{User} &The back was heavily damaged, and my car is undrivable. \\
    \textcolor{red}{Agent} &Was your car towed from the scene? \\
    \textcolor{blue}{User} &Yes, it was. Almost all of them had to be. \\
    \textcolor{red}{Agent} &Did the police come to the scene of the accident? \\
    \textcolor{blue}{User} &Yes, they did. They took statements from witnesses, and they also created an accident report that documented all involved parties' details. \\
    \textcolor{red}{Agent} &Great, do you happen to have the police report number and the name of the police department? \\
    \textcolor{blue}{User} &Yes, I have them right here. The police report number is 12345678, and it was the Philadelphia Police Department. \\
    \textcolor{red}{Agent} &Thank you, Joshua. Was anyone cited or received a ticket at the scene? \\
    \textcolor{blue}{User} &No, the police report stated that no party was cited. \\
    \textcolor{red}{Agent} &Okay, thank you for letting me know. Can you describe the traffic conditions at the time of the accident? \\
    \textcolor{blue}{User} &Traffic was flowing smoothly in the three-lane road. There was a car that stopped in the lane to my right, and the car behind them swerved into my lane. \\
    \textcolor{red}{Agent} &Did you notice any traffic signals or signs that may have contributed to the accident? \\
    \textcolor{blue}{User} &No, there weren't any traffic signals or signs at all in the tunnel. \\
    \textcolor{red}{Agent} &I see. Can you describe your car's make and model? What year was it made? And what color was it? \\
    \textcolor{blue}{User} &It's a white sedan, a 2018 Honda Accord. \\
    \textcolor{red}{Agent} &Thank you for that information, Joshua. Were there any witnesses to the accident? \\
    \textcolor{blue}{User} &Yes, there were several people who saw the accident happen. Some good Samaritans helped me after the accident and called 911. \\
    \textcolor{red}{Agent} &That's good to hear. Now, can you tell me about your injuries? What kind of medical treatment did you receive? \\
    \textcolor{blue}{User} &I dislocated my shoulder. They performed a CT scan at the hospital to ensure that there were no internal injuries. \\
    \textcolor{red}{Agent} &One more thing, Joshua. Can you remind me of the exact date and time of the accident? \\
    \textcolor{blue}{User} &It was on Monday morning, around 8:30 am. \\
    \textcolor{red}{Agent} &Okay, just to confirm, that would be the 22nd, correct? \\
    \textcolor{blue}{User} &Oh, wait. I think I may have remembered it wrong. It was actually last Tuesday. \\
    \textcolor{red}{Agent} &Thank you for clarifying the date, Joshua. Can you also tell me how fast were you driving when the accident occurred? \\
    \textcolor{blue}{User} &I was driving around 35 miles per hour. \\
    \textcolor{red}{Agent} &Thank you for that information, Joshua. Do you have the contact information for any of the other drivers? \\
    \textcolor{blue}{User} &Yes, I got Steve Woods' phone number. She was driving the yellow Dodge Ram that hit my car. \\
    \textcolor{red}{Agent} &Joshua, can you confirm how the accident occurred from your perspective? I'm a little unclear on some of the details. \\
    \textcolor{blue}{User} &Sure, I was driving in the middle lane and noticed a stopped car in the other lane. The yellow Dodge Ram swerved into my lane to avoid it. I couldn't change lanes because the other lane was bumper to bumper. The Dodge Ram hit my back bumper, bounced off, and hit the car on the right side of me. After that, several cars collided with each other. My car spun around and hit the tunnel wall, damaging the back. \\
    \textcolor{red}{Agent} &Thank you for explaining that, Joshua. Just to confirm, were there any traffic cameras at the scene that may have captured the accident? \\
    \textcolor{blue}{User} &No, I don't believe there were any traffic cameras. \\
    \textcolor{red}{Agent} &Thank you for providing all of that information, Joshua. Can I ask you to confirm some information about yourself? \\
    \textcolor{blue}{User} &Sure, go ahead. \\
    \textcolor{red}{Agent} &Can I have your full name? \\
    \textcolor{blue}{User} &Joshua Walters. That's spelled W-A-L-T-E-R-S. \\
    \textcolor{red}{Agent} &And can I have you provide your phone number and email address please? \\
    \textcolor{blue}{User} &Yes, my phone number is 555-1234, and my address is 123 Main Street. \\
    \textcolor{red}{Agent} &Great, thank you for providing that information. Now it looks like your insurance coverage includes liability insurance, collision coverage, rental car coverage, and gap insurance. \\
    \textcolor{blue}{User} &Okay. \\
    \textcolor{red}{Agent} &Do you understand how each of those coverages works? \\
    \textcolor{blue}{User} &Not really, can you explain it to me? \\
    \textcolor{red}{Agent} &Sure, liability insurance covers any damages or injuries that you may cause to another person or their property. Collision coverage pays for damages to your own vehicle if you're in an accident. Rental car coverage pays for a rental car while your vehicle is being repaired. And gap insurance covers the difference between the actual cash value of your car and the balance remaining on your car loan if your car is totaled. \\
    \textcolor{blue}{User} &Okay, I think I understand now. Thank you for explaining that to me. \\
    \textcolor{red}{Agent} &Do you have any other questions for me, Joshua? \\
    \textcolor{blue}{User} &No, I don't. Thank you for your help. \\
    \textcolor{red}{Agent} &Before we end this call, do you happen to have any other details about the accident that you would like to add? \\
    \textcolor{blue}{User} &No, I think that covers everything. \\
    \textcolor{red}{Agent} &Okay, thank you for providing all the information, Joshua. We will take care of everything from here. \\
    \textcolor{blue}{User} &Thank you so much for your help. \\
    \textcolor{red}{Agent} &Of course! And please don't hesitate to reach out if you have any questions or concerns. \\
    \textcolor{blue}{User} &I won't. Thanks again. \\\bottomrule
    \end{tabular}
    }
    \caption{Sample \syntaic dialogue 2.}
    \label{tab:sample_syntaic_ex2}
\end{table*}

%% file: tables/syntaic_examples/ex3.tex
\begin{table*}[ht]
    \centering
    \footnotesize
    \resizebox{0.98\linewidth}{!}{
    \begin{tabular}{rp{0.95\linewidth}}
    \toprule
    \textcolor{red}{Agent} &Thank you for calling DialGen Insurance! This is William. How may I help you today? \\
    \textcolor{blue}{User} &Hello. This is Roberta. I am calling for a car accident. \\
    \textcolor{red}{Agent} &I am sorry to hear that, Roberta. Before we get started, can I get your permission to record this call for quality and training purposes? \\
    \textcolor{blue}{User} &Yes, you can record the call. \\
    \textcolor{red}{Agent} &Thank you. May I have your full name and policy number, please? \\
    \textcolor{blue}{User} &My name is Roberta Oliver, and my policy number is XXXXXXXX. \\
    \textcolor{red}{Agent} &Got it. Can you please describe the accident to me? \\
    \textcolor{blue}{User} &Sure, I was driving my gold rideshare car on the highway after having dinner at a restaurant. I was changing lanes when Shane Scott's gray Chevrolet Camaro cut me off, and I had to slam on my brakes to avoid collision. Unfortunately, it was too late, and Shane's car crashed into mine. \\
    \textcolor{red}{Agent} &Okay, thank you for the detailed account. Were there any witnesses who saw this happen? Any traffic controls? \\
    \textcolor{blue}{User} &No, I didn't see any traffic controls around. I'm not sure about any witnesses. Oh, I guess there were the passengers in Shane's car, but they were too shaken up to give their statements to the police. \\
    \textcolor{red}{Agent} &Alright. How many passengers were in each car? \\
    \textcolor{blue}{User} &Shane had three passengers in her car. I was alone in mine. \\
    \textcolor{red}{Agent} &Thank you for that information, Roberta. Can you provide me with the location details of the accident as well as the date and time it occurred? \\
    \textcolor{blue}{User} &It was May 15th at around 4 in the afternoon. The accident happened on the highway near exit 45B. \\
    \textcolor{red}{Agent} &Thank you for sharing that information, Roberta. I forgot to ask earlier, what year is your car? \\
    \textcolor{blue}{User} &My car is a 2012 model. \\
    \textcolor{red}{Agent} &Great, thanks for letting me know. Can you describe the traffic conditions at the time of the accident? \\
    \textcolor{blue}{User} &It was a beautiful day, and the traffic on the highway was moving at a steady pace. There were four lanes, and we were both in the second lane from the left. \\
    \textcolor{red}{Agent} &Alright, I see. Before we proceed further, I want to let you know that I understand how stressful this situation can be. I want you to know that I am here to guide you through the process and make everything as clear and easy as possible. How are you feeling? \\
    \textcolor{blue}{User} &Honestly, I'm feeling pretty overwhelmed right now. My head has been hurting since the accident, and I'm worried about how much this is all going to cost. \\
    \textcolor{red}{Agent} &That's perfectly understandable, Roberta. Just take a deep breath and try to relax. It's good that you're taking steps towards resolving this by calling us today. Let's move forward together, okay? \\
    \textcolor{blue}{User} &Okay, thank you. \\
    \textcolor{red}{Agent} &Now you mentioned your head has been hurting since the accident. Did you injure your head during the crash? \\
    \textcolor{blue}{User} &Yeah, I hit my head on the steering wheel. Since then, I've been having constant headaches. It's been really difficult to focus on everyday tasks. \\
    \textcolor{red}{Agent} &I'm sorry to hear that. Have you seen a doctor yet? \\
    \textcolor{blue}{User} &Yes, I went to the hospital after the accident. They gave me a CT scan which revealed that I had a minor concussion. \\
    \textcolor{red}{Agent} &I'm sorry to hear that. Did they prescribe any treatment or medication? \\
    \textcolor{blue}{User} &Not really, other than rest and avoiding physical activities. They okayed me to go back home immediately, but I needed to have my husband check on me every few hours to make sure everything was fine that first night. \\
    \textcolor{red}{Agent} &Have you been back to the hospital since to follow up on the headaches? \\
    \textcolor{blue}{User} &No, but I did call my doctor to ask her about it. She said that headaches are normal for the first couple of months after a concussion, but to go back if they get worse. \\
    \textcolor{red}{Agent} &I see. Thank you for telling me that, Roberta, and I hope the headaches get better soon. Just a few more questions if you'll bear with me. Can you tell me which part of your car was damaged in the accident? \\
    \textcolor{blue}{User} &The front left side of my car was damaged. The back right side of Shane's car as well. \\
    \textcolor{red}{Agent} &Thank you for that information. Now I understand that it can be frustrating when there are no witnesses to corroborate your story. However, do you have any evidence of the accident? Perhaps photos of the damage or the police report? \\
    \textcolor{blue}{User} &Yes, the police came to file a report. I have a copy of it at home. I also took some photos of the damage to my car and Shane's car. \\
    \textcolor{red}{Agent} &Great, that will certainly help. Can you please send those photos over to our team? I can provide you with an email address where you can send them. \\
    \textcolor{blue}{User} &Sure, that would be helpful. What's the email address? \\
    \textcolor{red}{Agent} &The email is claims@DialGen Insurance.com. Please put your full name and policy number in the subject line and attach the photos in the email body. \\
    \textcolor{blue}{User} &Okay, thanks. I will send them over as soon as possible. \\
    \textcolor{red}{Agent} &Perfect. Is there anything else I can assist you with today, Roberta? \\
    \textcolor{blue}{User} &Yes, I was wondering about the insurance claim process. How long does it usually take to get a resolution? \\
    \textcolor{red}{Agent} &It depends on a few factors, such as the complexity of the case and how much evidence we have. Our team will carefully review your claim and reach out to you within a few business days with a resolution. \\
    \textcolor{blue}{User} &Okay, that's good to know. And what about rental cars or any other expenses related to the accident? \\
    \textcolor{red}{Agent} &We can certainly help you out with that if you need it. Our team can set up rental cars if necessary, and we will do everything we can to make sure you're not paying out of pocket for any expenses related to the accident. Will you be needing a rental car? \\
    \textcolor{blue}{User} &No, I don't think so. \\
    \textcolor{red}{Agent} &Alright, no problem. If you do end up needing a rental car, feel free to let us know. We're here to help in any way we can. \\
    \textcolor{blue}{User} &Thanks, I appreciate it. \\
    \textcolor{red}{Agent} &Of course, Roberta. Is there anything else I can assist you with today? \\
    \textcolor{blue}{User} &No, that's all for now. Thanks for your help, William. \\
    \textcolor{red}{Agent} &It was my pleasure, Roberta. Take care and have a great day! \\
    \textcolor{blue}{User} &You too. \\
    \bottomrule
    \end{tabular}
    }
    \caption{Sample \syntaic dialogue 3.}
    \label{tab:sample_syntaic_ex3}
\end{table*}